\pdfoutput=1
\documentclass[11pt]{article}
\usepackage[final]{acl}
\usepackage{times}
\usepackage{latexsym}
\usepackage[T1]{fontenc}
\usepackage[utf8]{inputenc}
\usepackage{microtype}
\usepackage{inconsolata}
\usepackage{graphicx}
\usepackage{algorithm}
\usepackage{algpseudocode}
\usepackage{subcaption} 
\usepackage{amsmath}
\usepackage{newfloat}
\usepackage{listings}
\usepackage{booktabs}
\usepackage{multirow}
\usepackage{url}
\usepackage{amsfonts}
\usepackage{nicefrac}
\usepackage{microtype}
\usepackage{xcolor}      % For colors (loaded once)
\usepackage{caption}
\usepackage{subcaption}
\usepackage{amssymb}
\usepackage{graphicx}
\usepackage{listings}
\usepackage{xpatch}
\usepackage{tikz}        % Loaded once
\usepackage{colortbl}
\usepackage{pifont}      % Loaded once
\usepackage{enumitem}    % Loaded once, also used for new dialogue environment
\usepackage{tabularx}
\usepackage{makecell}
\usepackage{arydshln}
\usepackage{subcaption} % 用于子图和子标题
\usepackage{longtable} % 用于创建可以跨页的长表格制
\usepackage{array} % 提供更灵活的列格式控制
\usepackage{xspace}
\usepackage{tabularx}

\title{DynamicNER: A Dynamic, Multilingual, and Fine-Grained Dataset for LLM-based Named Entity Recognition}

\author{
  Hanjun Luo$^{1} \thanks{ \url{hl6266@nyu.edu}}$,
  Yingbin Jin$^{3}$,
  Yiran Wang$^{5}$,
  Xinfeng Li$^{4}$,
  Tong Shang$^{6}$,
  Xuecheng Liu$^{2}$,\\
  \textbf{Ruizhe Chen}$^{2}$,
  \textbf{Kun Wang}  $^{4}$,
  \textbf{Hanan Salam}$^{1}$,
  \textbf{Qingsong Wen}$^{7}$,
  \textbf{Zuozhu Liu}$^{2} \thanks{Corresponding Author, \url{zuozhuliu@intl.zju.edu.cn}}$\\
  $^1$New York University Abu Dhabi, $^2$Zhejiang University,\\
  $^3$The Hong Kong Polytechnic University, $^4$Nanyang Technology University,\\
  $^5$University of Electronic Science and Technology of China,\\
  $^6$Texas A\&M University, $^7$Squirrel AI
}

\begin{document}
\maketitle
\begin{abstract}
The advancements of Large Language Models (LLMs) have spurred a growing interest in their application to Named Entity Recognition (NER) methods. However, existing datasets are primarily designed for traditional machine learning methods and are inadequate for LLM-based methods, in terms of corpus selection and overall dataset design logic. Moreover, the prevalent fixed and relatively coarse-grained entity categorization in existing datasets fails to adequately assess the superior generalization and contextual understanding capabilities of LLM-based methods, thereby hindering a comprehensive demonstration of their broad application prospects. To address these limitations, we propose DynamicNER, the first NER dataset designed for LLM-based methods with dynamic categorization, introducing various entity types and entity type lists for the same entity in different context, leveraging the generalization of LLM-based NER better. The dataset is also multilingual and multi-granular, covering \textbf{8} languages and \textbf{155} entity types, with corpora spanning a diverse range of domains. Furthermore, we introduce CascadeNER, a novel NER method based on a two-stage strategy and lightweight LLMs, achieving higher accuracy on fine-grained tasks while requiring fewer computational resources. Experiments show that DynamicNER serves as a robust and effective benchmark for LLM-based NER methods. Furthermore, we also conduct analysis for traditional methods and LLM-based methods on our dataset. Our code and dataset are openly available at \url{https://github.com/Astarojth/DynamicNER}.
\end{abstract}

\section{Introduction}
Recent advances in Large Language Models (LLMs) have transformed the landscape of NLP \citep{naveed2023comprehensive}. Among the impacted tasks, Named Entity Recognition (NER) has seen notable methodological shifts \cite{xie2023empirical}. Leveraging LLMs' strong generalization and contextual understanding capabilities, existing LLM-based methods \cite{shao2023prompt,li2023far} show superior performance compared to traditional machine learning (ML) methods \cite{wang2020automated,yan2021unified,curran2003language} in low-resource, multilingual, or few-/zero-shot settings. As a result, LLM-based NER has garnered growing interest in these settings \cite{xiao2024llm}, offering a promising path toward more scalable and adaptable information retrieval.

Despite recent progress, there are no existing NER datasets specifically optimized for the characteristics of LLMs, thereby limiting both their effective evaluation and the development of optimized methods. Existing NER datasets employ static categorization with a fixed set of entity types, preventing the  evaluation of LLMs' ability to generalize to novel entity types and varying levels of granularity, especially in few-shot or zero-shot settings. Moreover, while some datasets address domain-specific corpora with specialized entity types \cite{kim2003genia,liu-etal-2021-crossner}, others target multi-grained classifications \cite{ding-etal-2021-nerd}, or multilingualism \cite{malmasi-etal-2022-multiconer}, no existing dataset simultaneously incorporates all three aspects. This fragmentation hinders comprehensive evaluation of LLM-based methods, which are particularly well-suited to handling such challenges. As a result, current datasets fall short in revealing performance differences between LLM-based methods, fail to capture their full potential and limitations, and ultimately impede the advancement of more effective NER solutions.

To address these gaps, we develop DynamicNER, the first NER dataset optimized for LLM-based methods and the first to support dynamic categorization. It employs multiple strategies to dynamically adjust entity labels, type lists, and granularity levels during annotation. This design simulates the complexity and uncertainty of entity types in real-world, general-purpose scenarios—a challenge that traditional ML methods with fixed entity categorization struggle to address, enabling a more rigorous evaluation of LLM-based NER methods' ability to generalize across diverse and evolving scenarios. We introduce cohesion and distribution balance metrics to guide the evaluation and optimization of the annotation process. The entire procedure is algorithmically automated and openly available, ensuring both reliability and reproducibility.

In addition, DynamicNER is a multilingual and multi-granular dataset, featuring \textbf{8} languages, \textbf{8} coarse-grained types, \textbf{31} medium-grained types, and \textbf{155} fine-grained types. The multilingual and fine-grained nature of DynamicNER not only provides necessary support for the dynamic categorization, but also places higher demands on the NER methods. Many existing benchmarks are no longer difficult enough to reveal the performance boundaries of advanced methods, and the challenging DynamicNER will help future researchers to explore performance frontiers. Its entity types and corpora span a wide range of professional domains, including science, computer engineering, medicine, history, and arts. These features offer an unprecedented level of semantic and linguistic coverage for complex NER evaluation.

Furthermore, our evaluation on DynamicNER reveals significant limitations in existing LLM-based methods, particularly when migrating to lightweight LLMs (models with 1.5B to 7B parameters) for local deployment. While approaches leveraging commercial models like ChatGPT \cite{brown2020language} achieve high performance, this reliance introduces practical challenges related to API costs and privacy risks. API-based usage is often prohibitively expensive for real-world NER applications, and privacy remains a critical concern \cite{zhang2024right,das2024security,deng2025raconteur}.

To address this issue, we propose CascadeNER, a universal and multilingual NER framework that achieves competitive performance with lightweight LLMs, comparable to existing LLM-based methods that rely on costly commercial models. CascadeNER employs a two-stage strategy by dividing NER as two in-context text generation sub-tasks, extraction and classification, instead of treating it as a traditional sequential labeling task. To reduce task complexity and better capture in-context dependencies, CascadeNER assigns each stage of the NER process--extraction and classification--to separate fine-tuned lightweight LLMs within a model cascading framework \cite{varshney2022model}. This modular architecture, combined with the integration of prior knowledge, enables effective multilingual performance in low-resource settings.

We evaluate a BERT-based \cite{devlin2018bert} supervised method, two LLM-based methods, and our proposed CascadeNER on DynamicNER. We also conduct evaluations of CascadeNER against existing methods on existing datasets. Results demonstrate that DynamicNER effectively evaluates the performance of LLM-based methods in low-resource and complex NER tasks, while CascadeNER outperforms existing LLM-based methods significantly with smaller models. Moreover, this work offers the first comprehensive comparison and analysis of existing LLM-based NER methods, with an emphasis on multilingual and fine-grained scenarios.

Our contributions are summarized as follows:
\begin{itemize}[leftmargin=*]
\item[\ding{224}] We develop DynamicNER, the first NER dataset optimized for LLM-based NER methods, featuring a novel \textbf{dynamic categorization} system. The dataset which supports \textbf{8} languages, \textbf{155} entity types, and \textbf{3} levels of granularity, enabling comprehensive evaluation across diverse linguistic and semantic settings.

\item[\ding{224}] We propose CascadeNER, a universal NER framework, which outperforms existing LLM-based methods using only lightweight LLMs and a two-stage strategy.

\item[\ding{224}] We conduct a comprehensive evaluation of LLM-based NER methods on our challenging dataset and identify key challenges and future directions for the field.
\end{itemize}

\section{Related Works}
\paragraph{Named Entity Recognition.}
\label{related}
NER is the task of identifying named entities in text and classifying them into predefined categories. Supervised methods, such as BiLSTM \cite{yu2020named} and BERT-MRC \cite{li2019unified}, currently dominate this task. They generally rely on large amounts of training data to achieve strong performance, which limits their application in low-resource scenarios. Some researchers apply LLMs to address this issue. GPT-NER \cite{wang2023gpt} employs the GPT-3 model and re-frames the task as single-entity labeling, supporting few-shot/zero-shot learning. It achieves comparable performance to supervised methods in traditional scenarios and excels in low-resource scenarios. PromptNER \cite{ashok2023promptner} achieves state-of-the-art (SOTA) accuracy in datasets with complex classification \cite{liu-etal-2021-crossner,ding-etal-2021-nerd} with GPT-4 and Chain-of-Thought (CoT) \cite{wei2022chain}, yet performs significantly worse than GPT-NER and supervised methods in classical NER datasets like CoNLL2003 \cite{tjong-kim-sang-de-meulder-2003-introduction}. Furthermore, several studies apply LLM-based NER in domain-specific tasks \cite{li2023far, shao2023prompt, keloth2024advancing}, focusing on science and medicine. Their performances surpass supervised methods in those domains, further highlighting the potential of LLMs in low-resource and complex NER tasks.

\begin{table}[H]
    \centering
    \setlength{\tabcolsep}{1mm}
    \resizebox{0.48\textwidth}{!}{
    \begin{tabular}{lcccc}
        \toprule
        \textbf{Dataset} & \textbf{\#Language} & \textbf{\#Coarse} & \textbf{\#Fine} & \textbf{Domain} \\
        \midrule
        CoNLL2002 & 2 & 4 & no & News \\
        CoNLL2003 & 2 & 4 & no & News \\
        ACE2005 & 2 & 7 & 41 & News \\
        OntoNotes 5.0 & 3 & 18 & no & General \\
        CrossNER & 1 & 9-17 & no & Multi Domain \\
        FEW-NERD & 1 & 8 & 66 & General \\
        PAN-X & 282 & 3 & no & General \\
        MultiCoNER & 11 & 6 & 33 & General \\
        I2B2 & 1 & 22 & no & Medical \\
        \midrule
        \textbf{\texttt{DynamicNER (ours)}} & \textbf{8} & \textbf{8} & \textbf{155} & \textbf{Multi Domain} \\
        \bottomrule
    \end{tabular}
    }
    \caption{Overview of NER datasets. Notably, DynamicNER covers a wide range of cross-domain categories, such as art, medicine, and biology, thus offering better generalization compared to other general datasets.}
    \label{tab:dataset_summary}
\end{table}
\paragraph{NER Datasets.}
There have been a considerable number of NER datasets in various domains \cite{tjong-kim-sang-2002-introduction,kim2003genia,doddington2004automatic,walker2006ace,weischedel2011ontonotes,pradhan2013ontonotes,derczynski2017results,katz2023neretrieve}. However, these existing datasets exhibit several limitations, making them unsuitable for LLM-based NER. Most previous multilingual NER datasets adopt coarse-grained classification, no longer meeting the fine-grained requirements of contemporary flat NER applications. Even existing fine-grained datasets demonstrate clear limitations in category coverage and granularity, falling short of being truly "universal." For instance, FewNERD, despite having 66 entity types, suffers from highly imbalanced data distribution, which affects its reliability for evaluating few-shot learning capabilities. Furthermore, current datasets fail to adequately address the generalization capabilities of LLMs, hindering the comprehensive training and evaluation of LLM-based NER methods. Table \ref{tab:dataset_summary} presents a simple comparison between DynamicNER and existing multilingual or fine-grained datasets.

\section{DynamicNER Dataset}
\vspace{-1em}
\begin{figure}[H]
    \centering
    \includegraphics[width=0.48\textwidth]{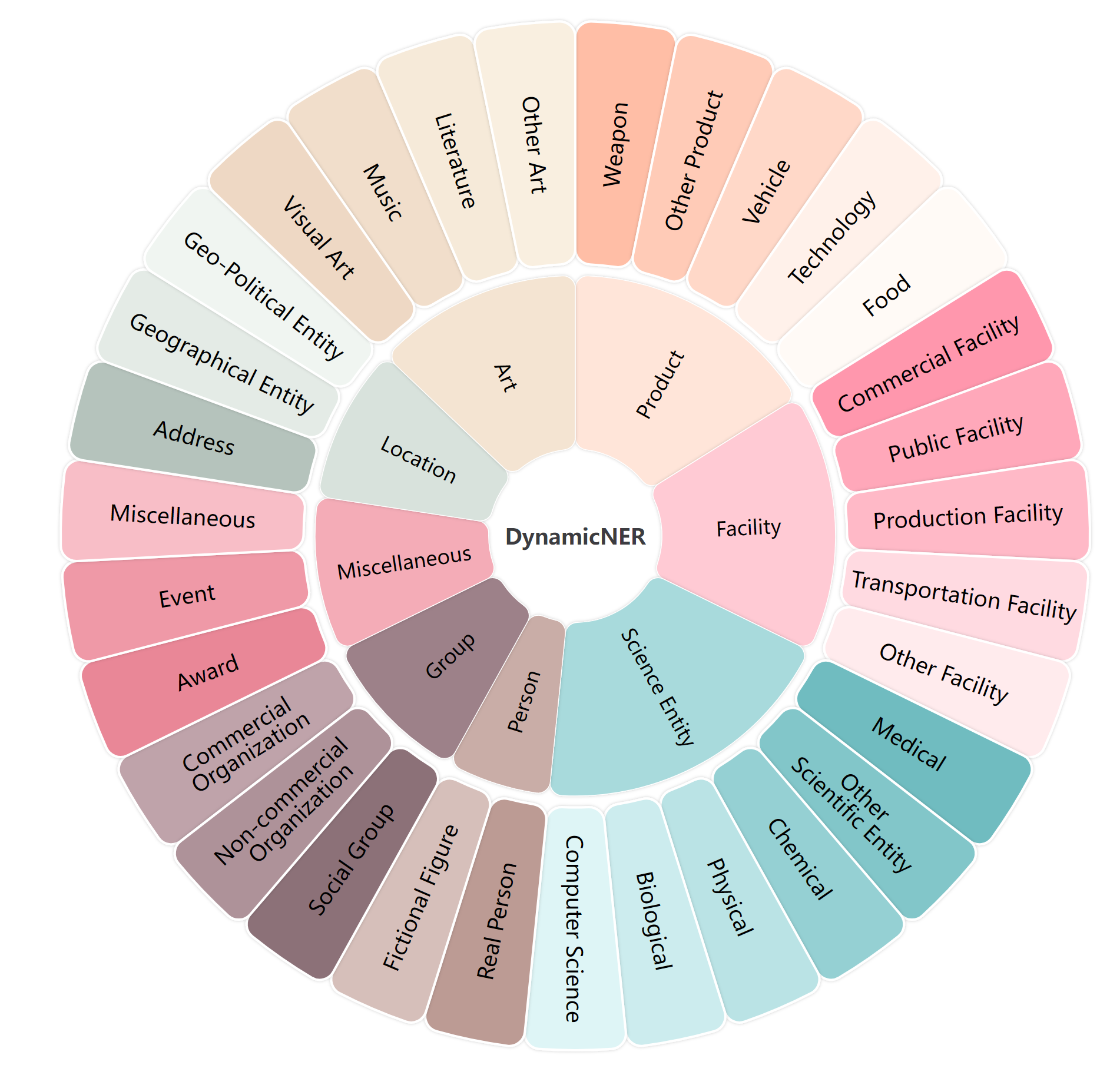}
    \vspace{-1em}
    \caption{The coarse-grained and medium-grained categories of DynamicNER. Detailed categories are provided in Appendix \ref{detailcategory}.}
    \label{anything_sta}
\end{figure}

 DynamicNER spans \textbf{8} languages: English, Chinese, Spanish, French, German, Japanese, Korean, and Russian. In terms of categorization, it is the first NER dataset with three-level granularity categorization, encompassing \textbf{8} coarse-grained types, \textbf{31} medium-grained types, and \textbf{155} fine-grained types, as shown in Figure \ref{anything_sta}. Like other NER datasets, DynamicNER is divided into train, dev, and test sets. Data volumes for different languages and parts shown in Appendix \ref{anything}. To develop DynamicNER, we first collect unlabeled corpus from Wikipedia. Then we manually extract sentences from corpora and conduct comprehensive manual annotation on the collected sentences, strictly following our 3-level, 155-type fine-grained entity schema. This initial annotated result, containing all entity types without any automated adjustments, is what we define as the \texttt{Base Version} of DynamicNER. It represents our most complete, static knowledge base. Subsequently, using the \texttt{Base Version} as input, we apply the automated dynamic categorization according to four metrics, generating the \texttt{Dynamic Version}. To ensure reproducibility and support future research on dynamic categorization, we provide the full code for the pipeline, including all settings and parameters in our repository.

\paragraph{Corpora Collection and Annotation.}
Wikipedia provides multilingual, domain-specific corpora with clear hierarchical and indexing systems, serving as a rich resource for our research. We utilize legal Wikipedia-API to filter and download corpora across different languages and categories, followed by manual selection and annotation of sentences. We particularly focus on corpora containing long texts and complex contexts, with a special emphasis on encyclopedic content from professional domains such as science, medical, arts, engineering, and law, to ensure DynamicNER's coverage of specialized domains. In addition, we have incorporated multilingual corpora from the social media platforms X and Weibo to ensure that our dataset also includes a substantial portion of colloquial language. After completing 50\% of the annotation process of each language, we annotate corpora from categories related to underrepresented entity types to achieve a balanced entity type distribution. For instance, when the entities of "Algorithm" are significantly less than others, we use more corpora from Computer Science category. Thus, DynamicNER ensures balanced entity distribution, and includes rare entities and emerging fields that are inadequate in existing datasets, ensuring comprehensive coverage across diverse domains. Details about the manual annotation process is provided in Appendix \ref{humananno}.

\begin{figure}[H]
    \centering
    \includegraphics[width=0.48\textwidth]{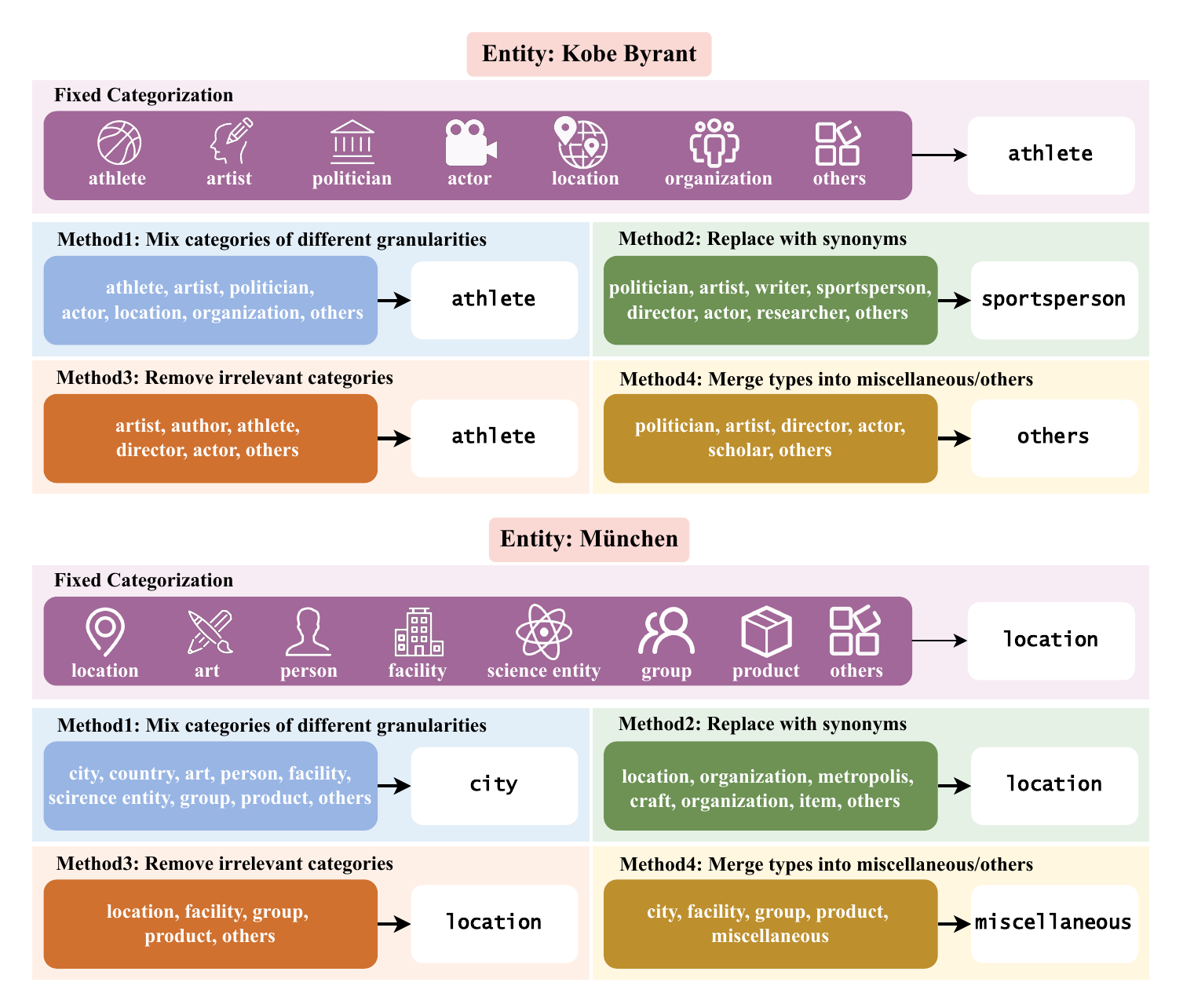}
    \vspace{-1em}
    \caption{Examples of dynamic categorization. Details of the four strategies are provided in Appendix \ref{strategies}.}
    \label{fig:dynamic_example}
\end{figure}

\paragraph{Dynamic Categorization.}
The dynamic version improves model generalization and reduces overfitting risk by dynamically adjusting entity labels and corresponding entity type lists during annotation, including \ding{202} mixing types of different granularities, \ding{203} replacing types with synonyms, \ding{204} using type lists without irrelevant types, and \ding{205} merging certain types into miscellaneous/others, as shown in Figure \ref{fig:dynamic_example}. This method addresses the mismatch between existing datasets and few-shot/zero-shot training needs, better simulating real-world scenarios, and is particularly critical for evaluating methods relying on complex prompt designs (e.g., CoT). Unlike traditional few-shot learning, some LLM-based methods only use few-shot demos to help the model understand the task or format, without requiring knowledge of entity types. They can perform NER across different datasets with fixed few-shot demos, resembling zero-shot NER. Research shows this method is more effective than typical zero-shot NER \cite{zhang2022automatic}. In methods that uses complex prompt designs like CoT to guide the reasoning, even few-shot CoT only conveys the CoT process rather than task-relevant knowledge, the performance of prompt-guided zero-shot CoT is significantly worse than few-shot CoT, making zero-shot restrictions inadequate for reflecting their true capability. However, in NER, models inevitably learn about entity types through few-shot demos, which limits generalization evaluations on fixed-category datasets. Our method significantly mitigates this limitation by varying entity types and lists, isolating the impact of prior type knowledge. Notably, as dynamic categorization is a subtractive method applied to a comprehensive and detailed categorization system, this method relies heavily on DynamicNER's fine-grained taxonomy, which includes 3 levels of granularity and 155 entity types. Consequently, this method may not be suitable for all datasets. Based on our empirical findings, we recommend that future researchers adopting this categorization method use datasets that feature at least two levels of granularity and over 20 fine-grained entity types.

\paragraph{Categorization Metrics.}
Random dynamic categorization not only exhibits poor reproducibility and explainability, but may also lead to data quality degradation. For training, inappropriate categorization may result in inconsistent learning objectives and overfitting risks \cite{ren2016label}. For evaluation, certain categories may experience imbalanced sample distribution and boundary ambiguity, reducing the comprehensiveness and consistency of evaluation \cite{obeidat2019description}. Thus, we design four metrics to regulate the dynamic categorization: cohesion, normalized entropy, Gini coefficient, and variation coefficient. The definition and calculation methods are provided in Appendix \ref{metric_detail}.

\begin{figure*}[t]
    \centering
    \includegraphics[width=1\textwidth]{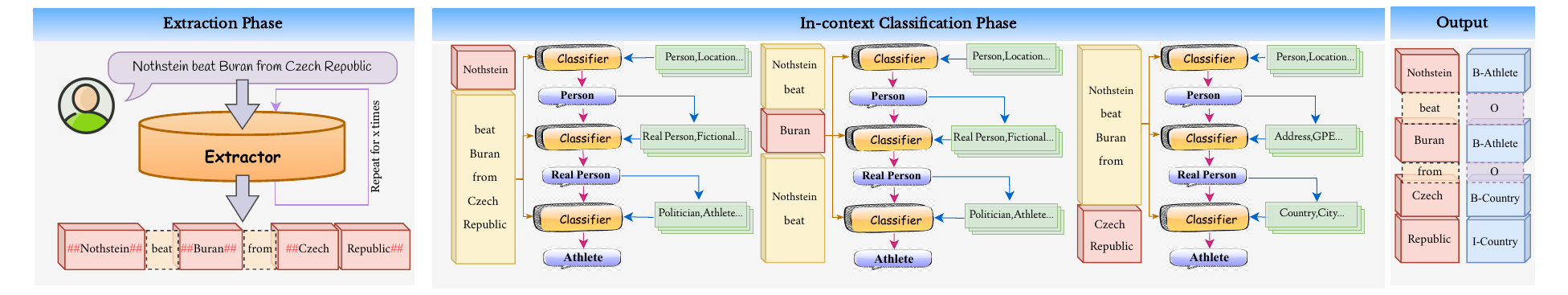}
    \vspace{-1em}
    \caption{Use a sentence and the multi-granularity categories of DynamicNER as the example. The extractor and classifier are the two different lightweight LLMs used in CascadeNER. Azure boxes represent the specific type list for the input of the classifier. Blue boxes represent the sentence input.}
    \label{fig:cascade_pipeline}
\end{figure*}
\begin{figure}[H]
    \centering
    \includegraphics[width=0.48\textwidth]{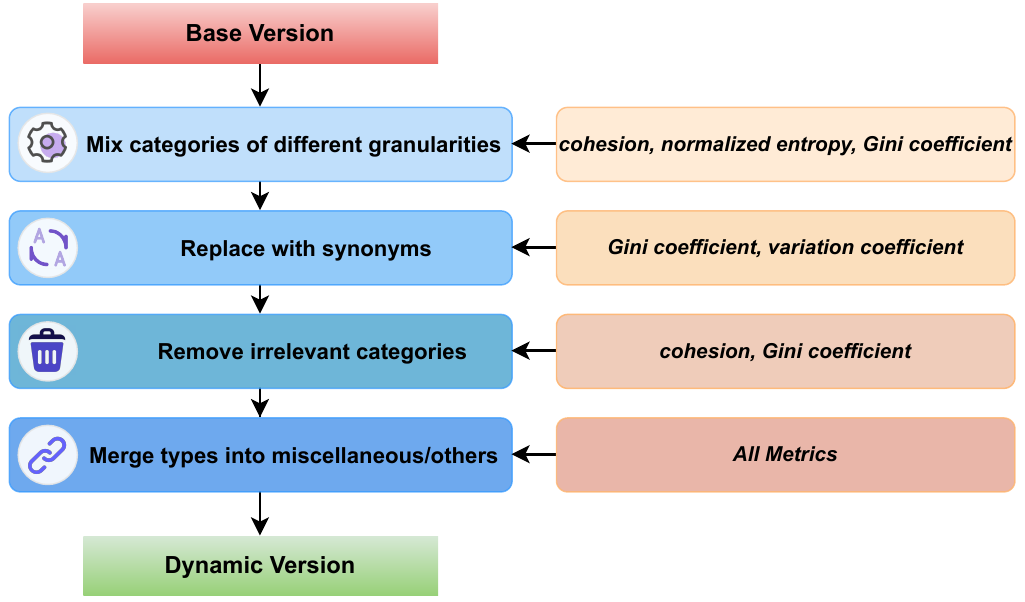}
    \vspace{-1em}
    \caption{Pipeline of dynamic categorization.}
    \label{fig:dynamic_pipeline}
\end{figure}

\paragraph{Categorization Process.}
The dynamic categorization process consists of 4 rounds of re-categorization, each sequentially corresponding to an adjustment method, and different metrics are employed in each round to guide the optimization. This hierarchical design enables each stage to focus on distinct data characteristics and optimization objectives, preventing interference between metrics while ensuring proper optimization direction, thus achieving a progressive optimization. We do not use all metrics in each evaluation, considering that certain metrics may have overlapping or conflicting effects at specific stages. For instance, normalized entropy and Gini coefficient both measure distribution uniformity, while improving cohesion may lead to more concentrated distribution and consequently lower entropy values. Figure \ref{fig:dynamic_pipeline} illustrates the metrics and methods corresponding to each round. Appendix \ref{metric_reason} explains the reasons of metric selection for each round.

\section{CascadeNER}
\subsection{Framework}
\paragraph{Background.}
Some existing supervised methods suggest that separating extraction and classification can improve NER performance as this two-stage strategy reduces the task complexity \cite{shen2021locate,wu2022propose}. However, these methods are limited by traditional models, failing to incorporate LLMs, and exhibit notable performance deficiencies that make them inferior to other methods treating NER as a single task. On the other hand, LLM-based methods demonstrate superior performance compared to traditional methods in Named Entity Extraction \cite{sancheti2024llm} and Text Classification \cite{gasparetto2022survey}, indicating the potential of two-stage in LLM-based NER.

\paragraph{Framework Design.}
We propose the framework to implement two-stage strategy in LLM-based NER. CascadeNER divides NER into two sequential, independently executed, generation-based sub-tasks. In the first sub-task, extraction, the model generate a sentence where all named entities are marked with identifiers and individually re-embeds each entity back into its context, resulting in sentences with identifiers at the number of entities. In the second sub-task, classification, the model receives sentences with identifiers and a list of entity types, and labels one entity at a time.

\paragraph{Model Cascading.}
To optimize performance while reducing computational resources, CascadeNER employs model cascading, where the extraction and classification sub-tasks are handled separately by two specialized fine-tuned LLMs. This structure allows each model to focus on its specific sub-task, maximizing performance on simpler, more specialized tasks. The architecture enables CascadeNER to be particularly suitable for lightweight LLMs, as each model only focus on a simplified task. Existing research shows that fine-tuned lightweight LLMs can achieve performance close to normal LLMs on specific simple tasks \cite{hu2024minicpm}. Through the implementation of two-stage strategy and model cascading, CascadeNER effectively leverages the advantages of lightweight LLMs in simple tasks, maintaining high accuracy while reducing computational resource usage.

\paragraph{Pipeline.}
A simplified pipeline of CascadeNER is shown in Figure \ref{fig:cascade_pipeline}. Upon receiving the input sentence, CascadeNER first processes the sentence by the extractor to mark all entities with identifiers, and re-embeds each entity back, resulting in sentences with identifiers around the named entities. These sentences are then individually fed into the classifier, which classifies each entity based on the context and the input type list. For multi-granularity data, CascadeNER allows a progressive strategy, significantly improving CasacdeNER's performance in accurate fine-grained classification. The detailed steps of extraction and classification are discussed in following sections. Examples about the prompts are provided in Appendix \ref{prompt_exp}.
 
\subsection{Extraction}
\label{extract_maintext}
\begin{figure}[H]
    \centering
    \includegraphics[width=0.48\textwidth]{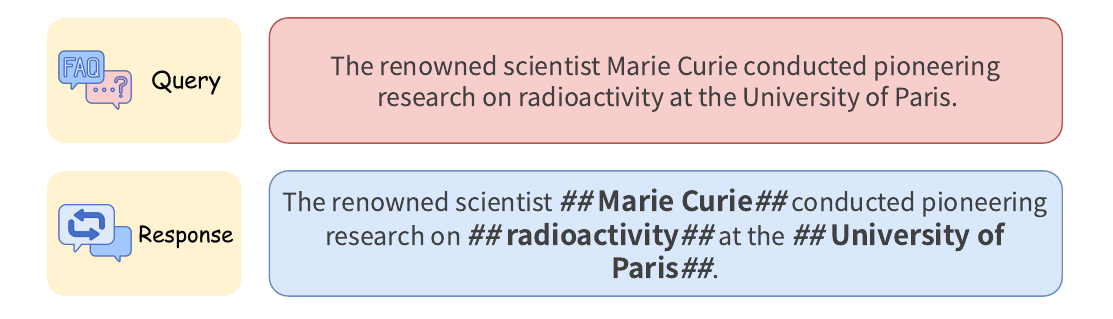}
    \vspace{-1em}
    \caption{Example prompts of extraction.}
    \label{fig:extract_example}
\end{figure}

\paragraph{Prompt Design.}
In the extraction sub-task, we utilize a generation-based extraction method, where special tokens "\#\#" are used to surround any entities identified in the sentence, regardless of the number of entities or their types. Figure \ref{fig:extract_example} shows an instance for the prompt and corresponding response.
This method, compared to conventional sequential extraction, avoids requiring LLMs to perform text alignment, thus reducing task complexity. Comparing similar methods \cite{wang2023gpt,hu2024improving}, CascadeNER's query contains only the sentence, without any task descriptions, demonstrations, or category information. The response exclusively uses "\#\#" as the identifiers, and all entities are extracted without specifying categories. CascadeNER achieves low-cost NER by using simple prompts and better generalization by treating all entities uniformly. A detailed comparison with existing methods and further advantages of our method are shown in Appendix \ref{extraction}.

\paragraph{Result Fusion.}
After conducting extensive experiments, we find that the extractor’s precision consistently exceeds recall, regardless of the model or dataset, indicating that while correct entities are effectively identified, there is a tendency for under-detection. To mitigate this issue, we introduce a union strategy in result fusion \cite{ganaie2022ensemble}, allowing multiple extraction for one sentence and taking the union of the results to maximize recall. For cases of entity nesting, where different extraction rounds produce overlapping or nested entities, we apply a length-first strategy, retaining the longer entity, as longer entities generally carry more semantic meaning \cite{nguyen2010nested}. For example, "Boston University" is semantically more accurate than "Boston" in the context of "She studies in Boston University". The formula of our strategy is shown below:
\begin{equation}
\small
E_{\text{final}} = \bigcup_{i=1}^{n} \left\{ \arg\max_{e \in E_i} \text{length}(e) \right\}
\end{equation}
where \( E_i \) is the set of extracted entities from the \(i\)-th extraction, \( n \) is the number of extraction rounds, \( E_{\text{overlap}} \) is the set of overlapping or nested entities, and \( \text{length}(e) \) is the length of entity \( e \). The effects of the number of extraction repetitions and other details are provided in Appendix \ref{fusion}. 

\subsection{Classification}
\begin{figure*}[t]
    \centering
    \includegraphics[width=0.95\textwidth]{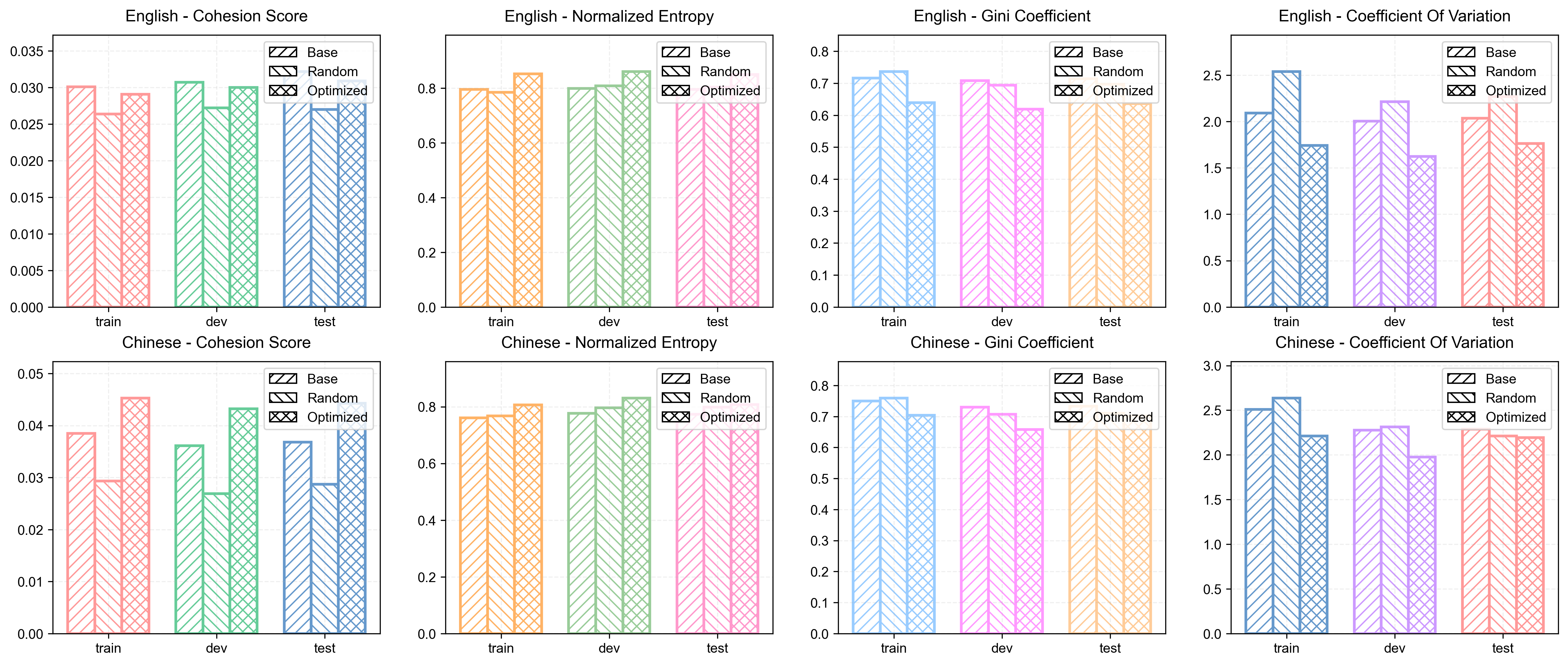}
    \caption{Quantitative categorization metric results for 3 versions DynamicNER in English and Chinese. Generally, higher cohesion and normalized entropy, or lower Gini coefficient and variation coefficient, indicate better quality.}
    \label{metrics_comparison_main}
\end{figure*}
\paragraph{Prompt Design.}
In the classification sub-task, we employ a generation-based in-context classification method, where we input the categories and the sentence with one entity surrounded by "\#\#", and require the classifier to generate the label for that entity. This method re-embeds the entity into the sentence for classification, which utilizes the self-attention architecture of LLMs for contextual information and improves accuracy compared to entity-level classification. Figure \ref{classif} is an example:
\begin{figure}[H]
    \centering
    \includegraphics[width=0.48\textwidth]{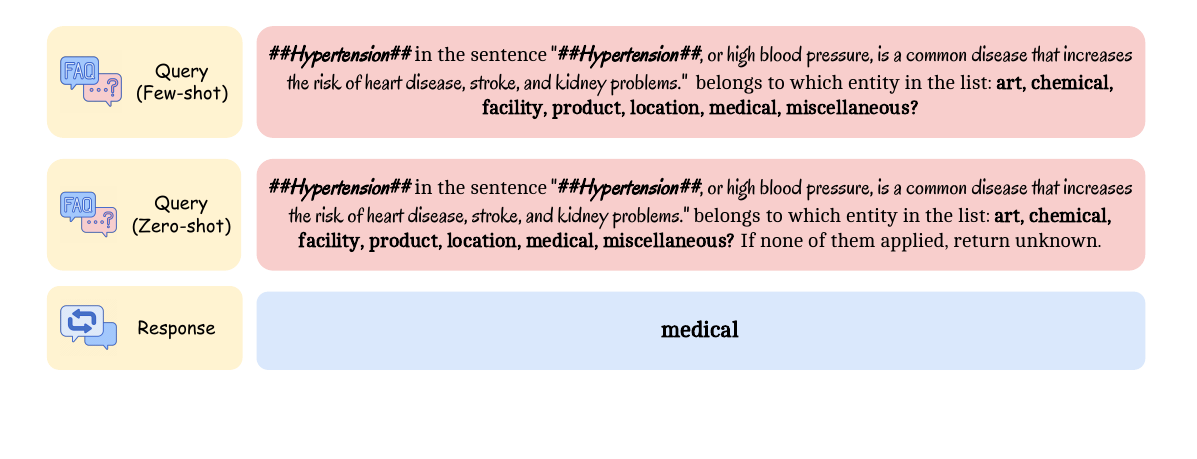}
    \caption{Example prompts of classification.}
    \label{classif}
\end{figure}
\vspace{-1em}

In zero-shot scenarios, we use a slightly different prompt. Due to differences in entity categorization across datasets, some entities in one dataset may be overlooked in others. We append the query with \texttt{If none of them applied, return unknown} to handle situations where the extracted entity cannot be classified into the provided categories, enhancing CascadeNER's generalization.

\paragraph{Multi-granularity.}
For multi-granularity data, we apply a progressive strategy. After obtaining the coarse-grained result, CascadeNER use the result to index the corresponding sub-categories and classify again, continuing this process until no further classification is possible:
\begin{equation}
\small
L_i^{\text{fine}} = f_{\text{fine-classify}}(L_i^{\text{coarse}}, \text{subcategories})
\end{equation}
where \( L_i^{\text{fine}} \) is the fine-grained label, \( L_i^{\text{coarse}} \) is the generated coarse-grained label, and \( \text{subcategories} \) are the subcategories under the coarse-grained.

\section{Experiment}
In this section, we first present the categorization metrics of DynamicNER before and after dynamic categorization, followed by a comparative analysis of existing methods and CascadeNER's performance on different versions of DynamicNER. We also conduct experiments of CascadeNER and baselines on existing datasets in Appendix \ref{cascadener}, and ablation studies in Appendix \ref{ablation}. In evaluations across existing datasets, CascadeNER, with base models fine-tuned using the dynamic version of DynamicNER, demonstrates consistent excellence in all datasets and achieves new SOTA performance in FewNERD and CrossNER. This confirms that DynamicNER not only provides exceptional effectiveness for evaluating LLM-based NER methods but also offers substantial value in training.

\begin{table*}[t]
\centering
\resizebox{1\textwidth}{!}{
\begin{tabular}{|l|cccccccc|cccccccc|}
\hline
\multirow{2}{*}{\textbf{Model}} & \multicolumn{8}{c|}{\textbf{Dynamic-Supervised}} & \multicolumn{8}{c|}{\textbf{Dynamic-Fewshot}} \\
\cline{2-17}
& \textbf{en} & \textbf{es} & \textbf{fr} & \textbf{ru} & \textbf{de} & \textbf{zh} & \textbf{ja} & \textbf{ko} & \textbf{en} & \textbf{es} & \textbf{fr} & \textbf{ru} & \textbf{de} & \textbf{zh} & \textbf{ja} & \textbf{ko} \\
\hline
\textbf{\texttt{GPT-NER-1.5B}} & 47.6 & 39.7 & 38.0 & 37.6 & 37.3 & 41.2 & 35.7 & 36.1 & 36.9 & 32.2 & 31.9 & 30.5 & 30.3 & 35.8 & 31.9 & 32.6 \\
\textbf{\texttt{GPT-NER-7B}} & 52.3 & 46.4 & 44.8 & 44.8 & 45.7 & 48.1 & 42.3 & 42.1 & 42.7 & 37.3 & 38.2 & 36.8 & 36.5 & 41.1 & 37.3 & 38.6 \\
\textbf{\texttt{GPT-NER-GPT}} & 60.6 & 57.3 & 56.5 & 55.6 & 55.9 & 58.4 & 54.9 & 53.8 & 49.2 & 46.9 & 47.5 & 47.2 & 47.0 & 48.9 & 47.7 & 48.3 \\
\hline
\textbf{\texttt{PromptNER-1.5B}} & 23.2 & 20.8 & 18.5 & 16.3 & 17.5 & 22.7 & 18.0 & 17.3 & 20.5 & 17.9 & 16.2 & 15.9 & 16.1 & 19.9 & 16.0 & 15.9 \\
\textbf{\texttt{PromptNER-7B}} & 44.3 & 35.8 & 33.2 & 32.5 & 31.9 & 40.4 & 37.4 & 35.6 & 39.8 & 33.2 & 32.1 & 31.8 & 31.5 & 37.8 & 35.6 & 34.5 \\
\textbf{\texttt{PromptNER-GPT}} & 53.0 & 50.5 & 51.2 & 47.9 & 50.2 & 52.3 & 48.7 & 48.5 & 49.4 & 48.5 & 47.1 & 46.6 & 46.0 & 47.4 & 44.1 & 44.0 \\
\hline
\textbf{\texttt{CascadeNER-1.5B}} & 62.8 & 55.7 & 52.8 & 51.1 & 48.8 & 58.9 & 54.1 & 52.7 & 49.7 & 44.1 & 44.0 & 43.4 & 42.9 & 48.5 & 43.1 & 43.8 \\
\textbf{\texttt{CascadeNER-7B}} & 68.2 & 61.5 & 55.3 & 52.9 & 51.4 & 64.5 & 58.8 & 55.3 & 55.7 & 49.9 & 49.7 & 46.5 & 46.1 & 52.9 & 50.2 & 50.0 \\
\textbf{\texttt{CascadeNER-GPT}} & \textbf{73.1} & \textbf{67.1} & \textbf{67.8} & \textbf{66.9} & \textbf{67.6} & \textbf{68.3} & \textbf{67.4} & \textbf{67.9} & \textbf{61.3} & \textbf{57.4} & \textbf{56.9} & \textbf{56.2} & \textbf{56.0} & \textbf{59.7} & \textbf{56.8} & \textbf{56.4} \\
\hline
\end{tabular}
}
\caption{The results of supervised learning with dynamic version and few-shot learning with dynamic version. G means GPT-NER, P means PromptNER, and C means CascadeNER. The results indicate that, due to its unprecedentedly detailed categorization and multilingual coverage, DynamicNER is an extremely challenging flat NER dataset, placing higher demands on methods' generalization capability.}
\label{dynamic-right}
\end{table*}

\subsection{Categorization Quality Evaluation}
To demonstrate that our dynamic categorization improves dataset generalization while maintaining dataset quality, we conduct comparative experiments across three versions of DynamicNER: the Base Version, a version with random parameters for dynamic categorization, and the optimized Dynamic Version. We still employ the 4 metrics for evaluating dataset quality, whose detailed definition are provided in Appendix \ref{metric_detail}. For reliability of the random version's results, we conduct five independent tests and use the average results. Due to space limitations, we only present results for English and Chinese in Figure \ref{metrics_comparison_main} here. Other quantitative results are provided in Appendix \ref{appendix_quality}.

\begin{table}[H]
    \centering
    \resizebox{0.48\textwidth}{!}{%
    \begin{tabular}{lcccccccc}
        \toprule
        \textbf{Language} & \textbf{en} & \textbf{es} & \textbf{fr} & \textbf{ru} & \textbf{de} & \textbf{zh} & \textbf{ja} & \textbf{kr} \\
        \midrule
        \textbf{\# Lists} & 725 & 455 & 501 & 377 & 465 & 786 & 553 & 478 \\
        \bottomrule
    \end{tabular}}
    \caption{The numbers of entity type lists of each language after dynamic categorization. In some scenarios, this can be equivalent to having 700+ distinct datasets.}
    \label{entity_types}
\end{table}
\vspace{-1em}
Experimental results demonstrate that our dynamic categorization significantly increases data diversity, as shown in Table \ref{entity_types}, while maintaining or improving dataset quality compared to the base version. The quality of the dynamic version also considerably surpasses the random version. These results comprehensively validate the reliability and effectiveness of our method.

\subsection{DynamicNER Experiment}

\paragraph{Baseline Selection.}
In our experiments for DynamicNER, we evaluate four NER methods: XLM-RoBERTa \cite{conneau2020unsupervised}, GPT-NER \cite{wang2023gpt}, PromptNER \cite{ashok2023promptner}, and our CascadeNER. XLM-RoBERTa is a famous BERT-based multilingual model widely used as a baseline in multilingual NER research \citet{malmasi-etal-2022-multiconer,fetahu2022dynamic}, thus being selected as our baseline representing supervised methods. GPT-NER and PromptNER are two major general LLM-based NER methods that achieve performance significantly superior to supervised methods in low-resource scenarios through sophisticated prompt design and powerful GPT models, as discussed in Section \ref{related} and Appendix \ref{extraction}.

\paragraph{Model Selection.}
Given the lack of existing lightweight LLM-based NER methods and controlled variable principles, we evaluate two LLM-based methods and CascadeNER using three LLMs: Qwen2.5-1.5B \cite{yang2024qwen2}, Qwen2.5-7B, and GPT-4o \cite{hurst2024gpt}. The lightweight LLMs of Qwen series perform exceptionally across benchmarks and gaining widespread recognition. According to \citet{huggingface2024review}, Qwen2.5-1.5B is the most downloaded open-source model in 2024. Therefore, we select Qwen2.5-1.5B and 7B to represent the current best-performing lightweight LLMs. GPT-4o is the most widely-used current general commercial LLM, and its previous versions are employed in GPT-NER and PromptNER, making it our choice. In CascadeNER, the extractor and the classifier use the same base model.

\paragraph{Implementation.}
For the supervised method, XLM-RoBERTa is trained and only evaluated with the base version of DynamicNER. As its fixed classification output layer corresponds to a predefined set of entity types and any modification to the entity type list necessitates full model retraining, it cannot be evaluated with the dynamic version. For LLM-based methods, we conduct experiments under three scenarios: supervised learning with base version, supervised learning with dynamic version, and few-shot learning with dynamic version. Training data for GPT-NER and CascadeNER is obtained through format conversion. For PromptNER, as its prompt involves complex designs such as CoT, we utilize LLM-generated prompts by GPT-4o using prompts from its paper as few-shot demonstrations and manually verified the prompts. The repetition count $i$ of CascadeNER for result fusion is set to 3. Potential data contamination are discussed in Appendix \ref{contamination}. The finetuning settings and hyperparameters of CascadeNER are provided in Appendix \ref{hyperparameter}.
\begin{table}[H]
\centering
\resizebox{0.48\textwidth}{!}{
\begin{tabular}{|l|cccccccc|}
\hline
\multirow{2}{*}{\textbf{Model}} & \multicolumn{8}{c|}{\textbf{Base-Supervised}} \\
\cline{2-9}
& \textbf{en} & \textbf{es} & \textbf{fr} & \textbf{ru} & \textbf{de} & \textbf{zh} & \textbf{ja} & \textbf{ko} \\
\hline
\textbf{\texttt{BERT}} & 41.9 & 33.5 & 29.1 & 23.4 & 32.9 & 29.2 & 27.2 & 28.6 \\
\hline
\textbf{\texttt{GPT-NER-1.5B}} & 50.2 & 43.5 & 40.4 & 39.8 & 39.3 & 44.1 & 38.9 & 38.7 \\
\textbf{\texttt{GPT-NER-7B}} & 55.1 & 48.2 & 47.2 & 44.0 & 48.1 & 50.9 & 44.8 & 44.5 \\
\textbf{\texttt{GPT-NER-GPT}} & 62.4 & 58.3 & 57.9 & 56.8 & 56.9 & 60.4 & 57.3 & 55.9 \\
\hline
\textbf{\texttt{PromptNER-1.5B}} & 21.6 & 18.6 & 17.1 & 14.9 & 15.8 & 20.7 & 16.4 & 15.9 \\
\textbf{\texttt{PromptNER-7B}} & 41.1 & 32.9 & 31.0 & 30.7 & 30.3 & 47.4 & 35.6 & 29.6 \\
\textbf{\texttt{PromptNER-GPT}} & 49.7 & 47.7 & 48.2 & 45.9 & 46.6 & 48.6 & 45.7 & 45.4 \\
\hline
\textbf{\texttt{CascadeNER-1.5B}} & 67.6 & 59.9 & 57.9 & 55.7 & 53.5 & 64.0 & 58.5 & 55.1 \\
\textbf{\texttt{CascadeNER-7B}} & 73.8 & 65.5 & 60.3 & 59.6 & 61.4 & 69.8 & 65.3 & 62.7 \\
\textbf{\texttt{CascadeNER-GPT}} & \textbf{77.1} & \textbf{71.7} & \textbf{69.9} & \textbf{70.3} & \textbf{70.8} & \textbf{74.3} & \textbf{72.4} & \textbf{70.9} \\
\hline
\end{tabular}
}
\caption{The results of supervised learning with base version. BERT represents XLM-RoBERTa. The supervised method XLM-RoBERTa performs terribly.}
\label{dynamic-left}
\end{table}
\paragraph{Results.}
As presented in Table \ref{dynamic-right} and \ref{dynamic-left}, CascadeNER achieves a significant advantage on DynamicNER, demonstrating its strong generalization. The supervised method XLM-RoBERTa performs terribly, as DynamicNER's low-resource characteristics make it more suitable for evaluating LLM-based methods. For LLM-based methods, the 3 methods show significant performance variations across different datasets and models. When using GPT-4o and transitioning from supervised to few-shot, PromptNER exhibits notably smaller performance degradation, partially reflecting the generalization advantages of reasoning-focused approaches. However, when migrating to lightweight LLMs, these methods show significantly larger performance drops compared to the other two methods. GPT-NER and CascadeNER demonstrate generally similar performance patterns, but GPT-NER shows more pronounced degradation when migrated to lightweight LLMs, while CascadeNER achieves a greater performance advantage on the dynamic version compared to the base version, validating the effectiveness of the two-stage strategy in the complex classification. CascadeNER also exhibits notable computational efficiency from two key perspectives. First, it is parameter-efficient, leveraging two 7B LLMs to achieve performance superior to baselines that rely on a 100B-level commercial model. Second, its two-stage design enables the use of concise prompts that drastically reduce token consumption compared to competing methods, as discussed in Appendix \ref{extraction}, as validated by our cost analysis in Appendix \ref{price}. This efficiency makes CascadeNER a powerful yet practical solution.

\subsection{Performance Across Granularities}

To investigate how different methods handle increasing classification complexity, we conduct an analysis across the three granularity levels of DynamicNER. We evaluate all LLM-based methods on the Base version (as the Dynamic version doesn't have clear granularity levels) for English and Chinese, with results presented in Table~\ref{tab:granularity_performance}.

The results lead to two key conclusions. First, as expected, the performance of all models declines as the number of entity types increases from coarse to fine, confirming that fine-grained classification is a significantly more challenging task. More importantly, the analysis reveals a stark difference in how the methods handle this increased complexity. Methods that rely on a single, complex prompt, such as PromptNER, suffer a dramatic performance collapse when scaled to the 155 categories. In contrast, our CascadeNER framework shows remarkable resilience, with a more graceful and controlled degradation in performance. This analysis empirically validates the core motivation for DynamicNER, demonstrating that a challenging, fine-grained benchmark is necessary to expose the limitations of existing methods and accurately evaluate the robustness of purpose-built frameworks.
\vspace{-1em}
\begin{table}[H]
\centering
\resizebox{0.48\textwidth}{!}{
\begin{tabular}{|l|ccc|ccc|}
\hline
\multirow{2}{*}{\textbf{Model}} & \multicolumn{3}{c|}{\textbf{English}} & \multicolumn{3}{c|}{\textbf{Chinese}} \\
\cline{2-7}
& Coarse & Medium & Fine & Coarse & Medium & Fine \\
\hline
\texttt{G-1.5B} & 59.3 & 55.0 & 50.2 & 59.7 & 54.5 & 44.1 \\
\texttt{G-GPT}  & 70.4 & 69.3 & 62.4 & 67.3 & 65.5 & 60.4 \\
\hline
\texttt{P-1.5B} & 58.5 & 49.2 & 21.6 & 57.7 & 48.8 & 20.7 \\
\texttt{P-GPT}  & 75.1 & 62.5 & 49.7 & 73.5 & 57.1 & 48.6 \\
\hline
\texttt{C-1.5B} & 82.7 & 78.8 & 67.6 & 81.2 & 75.1 & 64.0 \\
\texttt{C-GPT}  & \textbf{90.6} & \textbf{86.4} & \textbf{77.1} & \textbf{86.5} & \textbf{80.3} & \textbf{74.3} \\
\hline
\end{tabular}}
\caption{F1 scores across granularities on the Base-Supervised setting.}
\label{tab:granularity_performance}
\end{table}
\vspace{-1em}
\section{Conclusion}
This paper introduces DynamicNER, a multi-lingual and multi-granular NER dataset optimized for LLM-based NER, including a human-annotated base version and a dynamic-categorized version. We develop the first dynamic categorization method in NER datasets for DynamicNER, enhancing its generalization while keeping data quality. We also propose CascadeNER, a powerful NER framework which is exceptionally suitable for lightweight LLMs and local deployment, outperforming current LLM-based methods. Moreover, we conduct comprehensive experiments and analyses on DynamicNER and discuss the advantage and future direction of LLM-based NER. More experiments and discussions are provided in Appendix \ref{cascadener} and \ref{extraction}. We hope that DynamicNER and CascadeNER will facilitate future research in LLM-based NER, revitalizing this classical NLP task.

\newpage
\section*{Limitations}
There are still some challenges in our work. Although CascadeNER is designed to be able to accommodate nested and discontinuous NER, we only conduct evaluation on CascadeNER about flat NER tasks. This limitation arises from the fact that the models in CascadeNER are pre-trained on the dynamic version of DynamicNER, and DynamicNER is a flat NER dataset. Our resources are insufficient to collect enough open-source data for this purpose, which lead to DynamicNER containing only flat NER labels, and thus constraining CascadeNER to flat NER. Furthermore, Due to resource constraints and our failure to find annotators proficient in other languages for manual annotation, DynamicNER currently supports only 8 languages, which somewhat restricts its applicability.

\section*{Acknowledgements}
This work is supported by the National Key R\&D Program of China (Grant No. 2024YFC3308304), the "Pioneer" and "Leading Goose" R\&D Program of Zhejiang (Grant no. 2025C01128), the National Natural Science Foundation of China (Grant No. 62476241), the Natural Science Foundation of Zhejiang Province, China (Grant No. LZ23F020008), the State Key Laboratory (SKL) of Biobased Transportation Fuel Technology, and the Zhejiang Zhejiang Key Laboratory of Medical Imaging Artificial Intelligence.

This work is supported in part by the NYUAD Center for Artificial Intelligence and Robotics, funded by Tamkeen under the NYUAD Research Institute Award CG010 and in part by the NYUAD Center for Interdisciplinary Data Science \& AI (CIDSAI), funded by Tamkeen under the NYUAD Research Institute Award CG016.

Finally, we are grateful to the three anonymous reviewers for their insightful feedback, which has significantly improved the quality of this manuscript.
\newpage
\bibliography{refer}
% \normalsize
\newpage
\appendix

\section{Dynamic Categorization Strategies}
\label{strategies}

This section elaborates on the four strategies used to transform the manually annotated Base Version of DynamicNER into the final Dynamic Version. This entire process is an algorithm-driven pipeline that systematically modifies the entity type lists and corresponding labels for each data instance to create a more challenging and realistic benchmark. The pipeline consists of four sequential stages. We detail the goal, a concrete example, and the execution process for each strategy below.

\paragraph{Method 1: Mixing Categories of Different Granularities}
\begin{itemize}
    \item \textbf{Goal:} To test a method's robustness and generalization capability when faced with entity type lists of varying specificity. This strategy simulates the real-world variability where the required level of detail for an entity can change depending on the context.
    \item \textbf{Example:}
        \begin{itemize}
            \item \textit{Sentence:} ``Ada Lovelace is considered the first computer programmer.''
            \item \textit{Original State:} The entity ``Ada Lovelace'' is labeled as \texttt{Person} from the coarse-grained list, e.g., \texttt{[Person, Location, Organization, Product, ...]}.
            \item \textit{After Dynamic Change:} The system replaces the coarse-grained type with its fine-grained children. The new, shorter type list could be \texttt{[Artist, Athlete, Scholar, Location, Organization, Product, ...]}, and the required label for ``Ada Lovelace'' becomes \texttt{Scholar}.
        \end{itemize}
\end{itemize}

\paragraph{Method 2: Replacing Types with Synonyms}
\begin{itemize}
    \item \textbf{Goal:} To evaluate a method's ability to understand semantic equivalence and not just rely on keyword matching between the entity context and the type list.
    \item \textbf{Example:}
        \begin{itemize}
            \item \textit{Sentence:} ``Bekele won a gold medal in the competition.''
            \item \textit{Original State:} The entity ``Bekele'' is labeled as \texttt{Athlete} from a type list such as \texttt{[..., Artist, Author, Athlete, ...]}.
            \item \textit{After Dynamic Change:} The system replaces the type name \texttt{Athlete} with a valid synonym. The new type list becomes \texttt{[..., Artist, Author, Sportsperson, ...]}, and the required label for ``Bekele'' is now \texttt{Sportsperson}.
        \end{itemize}
\end{itemize}

\paragraph{Method 3: Removing Irrelevant Categories}
\begin{itemize}
    \item \textbf{Goal:} To simulate a more realistic, focused NER task where the type list is context-specific, and to test the model's ability to avoid hallucinating labels that are not in the provided list.
    \item \textbf{Example:}
        \begin{itemize}
            \item \textit{Sentence:} ``In the luxurious suite of The Ritz Carlton Hotel, a traveler is enjoying the beautiful view.''
            \item \textit{Original State:} The entity is ``The Ritz Carlton Hotel'', and the type list contains all fine-grained types of Commercial Facility, including many irrelevant ones like \texttt{Restaurant} and \texttt{Bank}.
            \item \textit{After Dynamic Change:} The system prunes unrelated categories. The new type list is \texttt{[Person, Location, Hotel, Market/Mall, Theater/Cinema, Other Commercial Facility]}.
        \end{itemize}
\end{itemize}

\paragraph{Method 4: Merging Types into Miscellaneous/Others}
\begin{itemize}
    \item \textbf{Goal:} To test the model's ability to correctly identify and classify infrequent or contextually less important entities into a general catch-all category, a common requirement in practical applications.
    \item \textbf{Example:}
        \begin{itemize}
            \item \textit{Sentence:} ``It was organized by Harry West and constituted a formal electoral pact.''
            \item \textit{Original State:} The entity ``Harry West'' is labeled as \texttt{Person} from the full type list.
            \item \textit{After Dynamic Change:} The new type list could be \texttt{[Location, Product, Group, Facility, Art, Miscellaneous]}, and the required label would become \texttt{Miscellaneous}.
        \end{itemize}
\end{itemize}
Through this four-stage automated pipeline, DynamicNER generates diverse and challenging evaluation scenarios that go beyond static categorization, enabling a more thorough evaluation of LLMs' true generalization and contextual understanding capabilities.

\section{Categorization Metric Definition}
\label{metric_detail}
\paragraph{Cohesion.}
Category cohesion score (cohesion) measures categorical semantic consistency by calculating the average semantic similarity between all entities within the same category. We employ the BERT-base \cite{devlin2018bert} model to extract semantic representations of entities, obtain embeddings of each entity, then computing cosine similarity between embeddings to derive cohesion. This metric ranges from [-1,1], where 1 indicates complete similarity and -1 indicates complete opposition. Typically, we perform category merging when cohesion exceeds 0.9. The formula is shown below:
\begin{equation}
    Cohesion = \frac{1}{n(n-1)} \sum_{i=1}^{n} \sum_{j=i+1}^{n} \cos(\mathbf{v}_i, \mathbf{v}_j)
\end{equation}
where $n$ is the number of entities in this category, $\mathbf{v}_i$ and $\mathbf{v}_j$ are the vector representations of the $i$-th and $j$-th entities encoded by BERT-base, and $\cos(\mathbf{v}_i, \mathbf{v}_j)$ represents the cosine similarity between two vectors.
\\
The detailed cosine similarity formula is shown below.
\begin{equation}
    \cos(\mathbf{v}_i, \mathbf{v}_j) = \frac{\mathbf{v}_i \cdot \mathbf{v}_j}{\|\mathbf{v}_i\| \|\mathbf{v}_j\|}
\end{equation}

\paragraph{Normalized Entropy.}
Normalized entropy measures the overall balance of category distribution. This metric is used for the influence of category quantity by calculating the information entropy of category frequency distribution and normalizing it to a score within the range [0,1]. A score of 1 indicates perfect balance, where all categories have equal sample sizes, while 0 indicates complete imbalance, where all samples are concentrated in a single category. When normalized entropy falls below 0.8, it indicates significant distributional imbalance and needs to be adjusted. The formula is shown below:
\begin{equation}
   H = -\frac{\sum_{i=1}^{n} p_i \log_2(p_i)}{\log_2(n)}
\end{equation}
where $n$ is the total number of categories, $p_i$ is the proportion of samples in the $i$-th category calculated as the number of samples in category $i$ divided by the total number of samples across all categories.

\paragraph{Gini Coefficient.}
The Gini Coefficient \cite{gini1921measurement} measures the degree of inequality in category distribution. Compared to normalized entropy, the Gini coefficient demonstrates higher sensitivity to distributional inequalities, performing better at identifying extreme imbalances where minority categories contain large sample proportions. For instance, when sample distributions exhibit extreme imbalances like [0.8, 0.1, 0.05, 0.05], the Gini coefficient provides stronger warning signals, while normalized entropy is more suitable for monitoring progressive imbalances such as [0.4, 0.3, 0.2, 0.1]. This metric also ranges from [0,1], where 0 indicates perfect balance and 1 indicates complete imbalance. A Gini coefficient exceeding 0.4 signals significant categorical inequality and requires distribution improvement. By using Gini coefficient and normalized entropy together, we achieve both sensitive detection of extreme imbalances and effective monitoring of overall distribution trends. The formula is shown below:
\begin{equation}
   G = \frac{n + 1 - 2\sum_{i=1}^{n}(n-i+1)p_i}{n}
\end{equation}
where $n$ is the total number of categories, $p_i$ is the proportion of samples in the $i$-th category after sorting proportions in ascending order $(p_1 \leq p_2 \leq ... \leq p_n)$.
\paragraph{Variation Coefficient.}
The Coefficient of Variation measures data dispersion by calculating the ratio of standard deviation to mean of category sample sizes. Its advantage is its scale independence, enabling comparisons across different scenarios. The coefficient ranges from 0 to positive infinity, where 0 indicates perfect balance and larger values indicate greater distributional imbalance. When the coefficient exceeds 0.5, it indicates significant fluctuation in sample sizes between categories, necessitating balance adjustments. The formula is shown below:
\begin{equation}
   CV = \frac{\sqrt{\frac{1}{n}\sum_{i=1}^{n}(x_i-\bar{x})^2}}{\bar{x}} = \frac{\sigma}{\mu}
\end{equation}
where $n$ is the total number of categories, $x_i$ is the number of samples in the $i$-th category, $\bar{x}$ is the mean number of samples across categories, $\sigma$ is the standard deviation of sample numbers, and $\mu$ is the mean.

\section{Categorization Metric Selection}
\label{metric_reason}
\paragraph{Mixing Types of Different Granularities.}
In this round of re-categorization, we use cohesion, normalized entropy, and Gini coefficient as metrics for optimization. Cohesion is employed to assess relationships between entity types, where close categorical relationships reduce the need for mixing to avoid creating unreasonable combinations. Meanwhile, normalized entropy and Gini coefficient are utilized to comprehensively measure distribution uniformity, where uneven distributions guide the system to perform additional merging for balance or category redistribution.
\paragraph{Replace with Synonyms.}
In this round of re-categorization, we use Gini coefficient and variation coefficient as metrics for optimization. We employ the variation coefficient to measure data dispersion, increasing synonym substitutions for increasing data convergence when dispersion is high. The Gini coefficient is used to guide system to reduce operations to prevent exacerbating imbalances when distributions are uneven. Cohesion is not used as synonym substitution does not alter hierarchical relationships between categories. Entropy is also given up because synonym substitution primarily focuses on linguistic variation rather than distributional changes.
\paragraph{Remove Irrelevant Types.}
In this round of re-categorization, we use cohesion and normalized entropy as metrics for optimization. We employ normalized entropy as a reference for controlling removal probability, ensuring that deletion operations do not result in overly concentrated distributions. Additionally, the system adjusts removal probability when cohesion is low, regulating relationships between categories. The variation coefficient is not used as this stage primarily focuses on option quantity rather than distribution characteristics, while the Gini coefficient is omitted since distribution balance has been addressed in the previous two stages, thus temporarily foregoing the Gini coefficient to prevent interference with other metrics.
\paragraph{Merge Types into Miscellaneous.}
In this round of re-categorization, we use all four metrics for optimization. As the final optimization stage, it requires consideration across all dimensions. We use all metrics for final fine-tuning to ensure overall data quality and avoid biases that might arise from single metric optimization.

\section{Detail of DynamicNER}
\label{anything}
The specific data volumes for each language are shown in Table \ref{any_stat}. It is important to note that for languages except English and Chinese, we partially use manually translated English corpora. This is necessary to balance category distribution, as some languages lack sufficient corpora in specific domains. For the colloquial portion of the corpora, the Chinese part is sourced from Weibo, while the data for all other languages originated from X platform. We provide conversion scripts that allow DynamicNER to be transformed into train, dev, and test sets with non-overlapping subsets based on coarse categories, making it easier to use for traditional few-shot learning methods.
\begin{table}[H]
\centering
\setlength{\tabcolsep}{1mm}
\resizebox{0.48\textwidth}{!}{
\begin{tabular}{lcccccc}
\toprule
\textbf{Language} & \textbf{\# Sentences} & \textbf{\# Tokens} & \textbf{\# Entities} & \textbf{\# Train} & \textbf{\# Dev} & \textbf{\# Test} \\
\midrule
\textbf{English}  & 1500 & 36.7k & 4664 & 300 & 300 & 900 \\
\textbf{Chinese}  & 1500 & 98.1k & 5198 & 300 & 300 & 900 \\
\textbf{Spanish}  & 1000 & 22.8k & 2454 & 197 & 201 & 602 \\
\textbf{French}   & 1000 & 24.1k & 2763 & 200 & 200 & 600 \\
\textbf{German}   & 1000 & 21.7k & 2800 & 200 & 197 & 603 \\
\textbf{Japanese} & 1000 & 81.7k & 3032 & 201 & 199 & 600 \\
\textbf{Korean}   & 1000 & 66.4k & 2401 & 202 & 200 & 598 \\
\textbf{Russian}  & 1000 & 18.5k & 2092 & 201 & 198 & 601 \\
\bottomrule
\end{tabular}
}
\caption{Statistics of DynamicNER across languages. We roughly follow a 1:1:3 ratio to divide the train, dev, and test sets, with slight adjustments based on the proportional distribution of entities within the corpus.}
\label{any_stat}
\end{table}
\vspace{-0.5em}
Additionally, two points about the Table \ref{any_stat} require clarification. First, as DynamicNER's design emphasizes the evaluation of generalization and low-resource learning capabilities, we set the test set capacity to the biggest one, rather than the train set. Second, for Chinese, Japanese, and Korean, due to linguistic characteristics where each character is treated as a token, the token count appears significantly higher, though the actual corpus volume is comparable to other languages.
\\
Although DynamicNER’s data volume is significantly smaller than existing multilingual NER datasets, it aligns with our primary goal: evaluating LLM-based NER in fine-grained, low-resource, and dynamic scenarios. In such contexts, data volume may not be decisive. We acknowledge that smaller data volumes may lead to higher variance and overfitting. To show the robustness of our work, we conduct additional experiments on existing datasets with the model finetuned on DynamicNER. As shown in Appendix \ref{cascadener}, especially Table \ref{english} \& \ref{multiling}, CascadeNER, trained on DynamicNER, demonstrates strong few-shot performance on other established benchmarks. This indicates that the models have learned robust, generalizable features, avoiding overfitting.

\section{More Experiment about CascadeNER}
\label{cascadener}

\subsection{CascadeNER Setting}
In the experiments of this section, CascadeNER always employs two Qwen2.5-7B base models, which are fine-tuned separately based on the corresponding part of the dynamic version of DynamicNER to obtain an extractor and a classifier. Potential data contamination about the fine-tuning is discussed in Appendix \ref{contamination}. We evaluate CascadeNER's performance in both few-shot and zero-shot scenarios, comparing it with supervised SOTAs and LLM-based baselines. For few-shot scenarios, the number of few-shot demonstrations is set to $3$, the same as the experiments on DynamicNER.

\subsection{Baselines}
For supervised methods, we adopt ACE+document-context by \citet{wang2020automated} (SOTA of CoNLL2003) and BERT-MRC+DSC by \citet{li2019dice} (SOTA of Ontonotes 5.0 \cite{pradhan2013ontonotes}) for English datasets, while XLM-RoBERTa \cite{conneau2020unsupervised} and GEMNET by \citet{meng-etal-2021-gemnet,fetahu2022dynamic} (SOTA of MultiCoNER) for multilingual datasets. For LLM-based methods, we adopt GPT-NER and PromptNER with GPT-4o.

\subsection{Dataset}
\begin{table*}[t]
    \centering
    \resizebox{\textwidth}{!}{%
    \begin{tabular}{lcccccccc}
        \toprule
        \textbf{Model} & \textbf{CoNLL2003} & \textbf{AI} & \textbf{Literature} & \textbf{Music} & \textbf{Politics} & \textbf{Science} & \textbf{FewNERD-8} & \textbf{FewNERD-66} \\
        \midrule
        \textbf{XLM-RoBERTa} & 92.3 & 59.0 & 65.9 & 72.1 & 70.8 & 66.9 & 80.5 & 64.1 \\
        \textbf{ACE+document-context} & \textbf{94.6} & 17.2 & 22.6 & 23.8 & 35.1 & 32.3 & 83.3 & 70.4 \\
        \textbf{BERT-MRC+DSC} & 93.5 & 63.2 & 67.8 & 74.5 & 76.1 & 68.7 & \textbf{86.7} & 74.1 \\
        \textbf{PromptNER} & 84.2 & 64.8 & 74.44 & \textbf{84.2} & 78.6 & 72.6 & 76.5 & 35.6 \\
        \textbf{GPT-NER} & 73.5 & 58.0 & 61.2 & 60.8 & 62.4 & 55.8 & 70.0 & 58.4 \\
        \midrule
        \textbf{CascadeNER (zero-shot)} & 88.2 & 68.9 & 71.7 & 79.3 & 80.5 & 73.6 & 73.4 & 67.0 \\
        \textbf{CascadeNER (few-shot)} & 92.8 & \textbf{75.8} & \textbf{75.2} & 83.2 & \textbf{82.4} & \textbf{77.1} & 84.5 & \textbf{75.9} \\
        \bottomrule
    \end{tabular}}
    \caption{F1 score of different models on CoNLL2003, CrossNER, and FewNERD.}
    \label{english}
\end{table*}

% You might need these packages in your preamble
% \usepackage{graphicx}
% \usepackage{float}
% \usepackage{amsmath}

\begin{table*}[t]
\centering
\resizebox{1\textwidth}{!}{%
\begin{tabular}{lcccccccccccccc}
\toprule
\multirow{2}{*}{\textbf{Model}} & \multicolumn{8}{c|}{\textbf{PAN-X}} & \multicolumn{6}{c}{\textbf{MultiCoNER}} \\
\cmidrule(lr){2-9} \cmidrule(lr){10-15}
& en & es & fr & ru & de & zh & ja & ko & en & es & ru & de & zh & ko \\
\midrule
\textbf{XLM-RoBERTa} & 88.1 & 86.5 & 85.4 & 86.3 & 83.1 & 78.3 & 75.6 & 82.0 & 58.9 & 54.8 & 55.9 & 60.6 & 62.6 & 52.0 \\
\textbf{GEMNET} & 90.5 & \textbf{91.1} & \textbf{87.6} & \textbf{87.4} & \textbf{86.6} & 81.5 & 80.8 & \textbf{85.5} & 84.3 & \textbf{85.3} & 78.7 & \textbf{89.5} & 83.2 & \textbf{85.7} \\
\textbf{PromptNER} & 81.7 & 79.6 & 73.5 & 73.8 & 71.9 & 72.1 & 70.8 & 73.5 & 79.5 & 75.6 & 76.5 & 67.6 & 70.8 & 72.4 \\
\textbf{GPT-NER} & 75.2 & 72.8 & 71.6 & 63.5 & 72.0 & 72.4 & 71.5 & 72.1 & 71.7 & 67.9 & 58.2 & 63.1 & 61.2 & 62.5 \\
\midrule
\textbf{CascadeNER (zero-shot)} & 87.8 & 85.0 & 83.2 & 80.7 & 77.4 & 78.7 & 74.7 & 72.0 & 71.9 & 71.5 & 71.2 & 63.5 & 70.3 & 69.8 \\
\textbf{CascadeNER (few-shot)} & \textbf{91.0} & 85.2 & 87.2 & 86.8 & 82.8 & \textbf{87.0} & \textbf{83.2} & 79.4 & \textbf{85.9} & 81.1 & \textbf{79.5} & 69.1 & \textbf{85.1} & 76.9 \\
\bottomrule
\end{tabular}}
\caption{F1 score of different models across languages on PAN-X and MultiCoNER.}
\label{multiling}
\end{table*}
\paragraph{Few-shot Data Sampling.}
In existing datasets, only CrossNER \cite{liu-etal-2021-crossner}, designed for low-resource scenarios, and FewNERD \cite{ding-etal-2021-nerd}, designed for few-shot scenarios, meet our requirements for evaluating CascadeNER in few-shot scenarios. However, relying solely on them is insufficient for comprehensively evaluating CascadeNER, particularly its multilingual NER performance. To address this, we develop a sampling algorithm to construct datasets for few-shot evaluation. Considering that basic random sampling cannot ensure a balanced category distribution, we employ a stratified sampling algorithm, which divides the dataset into strata based on the labels. Each stratum corresponds to a distinct entity type, and we ensure an relatively equal number of samples per category by drawing from these strata, thereby maintaining balance across categories in the results. The size for each stratum is calculated with the formula:
\begin{equation}
\small
s_i = \min\left(\left\lfloor \frac{S}{m} \right\rfloor, n_i\right)
\end{equation}
where $N$ is the total number of labels in the dataset, $S$ is the total sample size, $n_i$ is the total number of labels with value $i$, $m$ is the number of categories, and $s_i$ is the number of labels from stratum $i$.
\paragraph{Dataset Selection.}
We conduct supplementary experiments on existing datasets including CoNLL2003 \cite{tjong-kim-sang-de-meulder-2003-introduction}, CrossNER \cite{liu-etal-2021-crossner}, FewNERD \cite{ding-etal-2021-nerd}, PAN-X \cite{pan-etal-2017-cross}, and MultiCoNER \cite{malmasi-etal-2022-multiconer}. Since we decide to use and share the formatted versions of these datasets in our repository to facilitate the test and use of CascadeNER, we only choose open-sourced datasets to avoid copyright issues. For evaluation metrics, we primarily use F1 score, as it is widely recognized as the most robust and effective metric for NER tasks \cite{li2020dice}. We detail below the reasons for selecting these datasets and their usage.

\paragraph{CoNLL2003.}
CoNLL2003 is the most widely used English NER dataset, featuring four types: PER, LOC, ORG, and MISC. Supervised methods achieve excellent F1 scores of 90\%-95\% on this dataset. We use this dataset to compare CascadeNER and other LLM-based methods with existing supervised SOTAs in classical scenarios.

\paragraph{CrossNER.}
CrossNER is a English cross-domain dataset primarily used to evaluate a model's cross-domain generalization and low-resource performance. It consists of five independent sub-datasets, each covering a specific domain (AI, Literature, Music, Politics, and Sciences) and containing 9-17 entity types. Since the train set for the datasets only contains 100-200 sentences, supervised methods underperform compared to LLM-based methods. We use this dataset to evaluate CascadeNER in cross-domain and low-resource scenarios. 

\paragraph{FewNERD.}
FewNERD is an English dataset designed to evaluate a model's ability to handle fine-grained entity recognition and few-shot learning, comprising 8 coarse-grained types and 66 fine-grained types. For supervised methods, FewNERD applies all 66 categories, challenging the models' classification abilities. For few-shot methods, we use the Intra-10way setting, where the train, dev, and test sets contain non-overlapping entity types. We utilize both the 8-category and 66-category settings to evaluate CascadeNER under varying levels of classification granularity.

\paragraph{MultiCoNER \& PAN-X.}
MultiCoNER and PAN-X are two widely used multilingual datasets. MultiCoNER covers 6 entity types and 11 languages, while PAN-X includes 3 entity types and 282 languages. We use 6 and 8 overlapping languages from MultiCoNER and PAN-X with DynamicNER to evaluate CascadeNER's multilingual capabilities. Notably, for the purpose of controlling variables, all methods requiring training are trained using multilingual joint training.

\subsection{Experimental Results}
 As shown in Table \ref{english} and \ref{multiling}, the results indicate that in low-resource scenarios, LLM-based methods achieve significantly better results. CascadeNER surpasses existing methods on CrossNER except Music and FewNERD, and PAN-X and MultiCoNER in some languages, achieving new SOTA performance and highlighting its exceptional generalization and capability to handle complex entity categorization. However, when handling NER tasks with ample training resources and simple classifications, LLM-based methods still lag behind existing methods, whether on the English-only CoNLL2003 or the multilingual PAN-X, indicating that supervised methods are still useful in some scenarios.

\section{Ablation Study}
\label{ablation}
\subsection{Result Fusion}
\label{fusion}
\begin{figure}[H]
    \centering
    \includegraphics[width=0.46\textwidth]{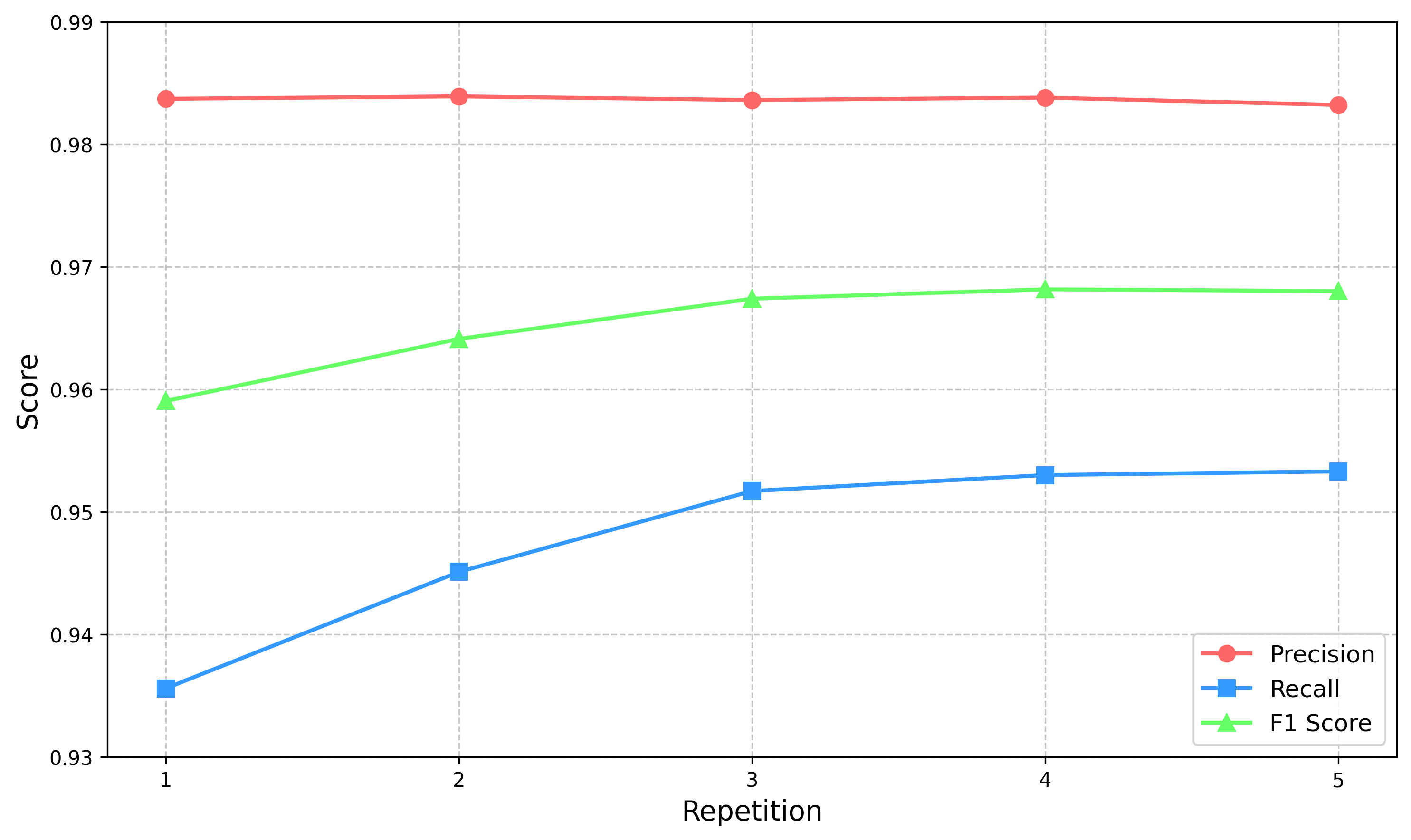}
    \caption{The curves showing visualized precision, recall, and F1 Score as a function of the number of repetitions, demonstrating how these metrics change with increasing repetition counts $k$. Both base models used in CascadeNER are Qwen2.5-7B.}
    \label{repeat}
\end{figure}

In Section \ref{extract_maintext}, we introduce our union strategy in result fusion to address the issue of extractor recall being significantly lower than precision, allowing multiple extractions for one sentence and taking the union of the results to maximize recall. For the problem of entity nesting, where different extraction rounds yield overlapping or nested entities, we adopt a length-first strategy, retaining the longer entity. Table \ref{recall} provides a example for the significantly low recall.
\begin{table}[H]
    \centering
    \setlength{\tabcolsep}{1mm}
    \resizebox{0.4\textwidth}{!}{
    \begin{tabular}{lccc}
        \toprule
        \textbf{Dataset} & \textbf{Precision} & \textbf{Recall} & \textbf{F1 Score} \\
        \midrule
        \textbf{CoNLL2003} & 98.4 & 93.6 & 95.9 \\
        \textbf{AI} & 98.7 & 88.0 & 93.1 \\
        \textbf{Literature} & 98.3 & 87.8 & 92.7 \\
        \textbf{Music} & 98.0 & 92.0 & 94.9 \\
        \textbf{Politics} & 97.5 & 90.0 & 93.6 \\
        \textbf{Science} & 98.2 & 85.9 & 91.6 \\
        \bottomrule
    \end{tabular}}
    \caption{Precision, recall, and F1 Score for CoNLL2003 and CrossNER. In this experiment, both base models used in CascadeNER are Qwen2.5-7B, and the results are obtained in zero-shot scenarios.}
    \label{recall}
\end{table}
\noindent
Figure \ref{repeat} presents the impact of increasing the number of extraction repetitions in zero-shot scenarios on CoNLL2003. The results show that our strategy can slightly improve recall with minimal impact on precision. Given the obvious margin effect after 3 repetitions, we ultimately select $3$ as the repetition count $k$ for other experiments. It is important to emphasize that even without repetition, CascadeNER still has a significant performance advantage.

\subsection{Context in Classification}
\label{context}
\begin{figure}[H]
    \centering
    \includegraphics[width=0.48\textwidth]{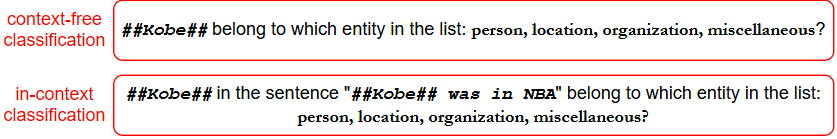}
    \caption{Example of the early context-free queries.}
    \label{contextfree}
\end{figure}

In the early stages of our research, the prompt used for classification contained only the entity itself without any context. Figure \ref{contextfree} provides an example comparing the two types of prompts. Although this method makes the prompt more concise, it lacks any contextual information. Our final in-context classification queries significantly improve classification accuracy, as shown in Table \ref{context_compare}.
\begin{table}[H]
    \centering
    \setlength{\tabcolsep}{1mm}
    \resizebox{0.46\textwidth}{!}{
    \begin{tabular}{lcc}
        \toprule
        \textbf{Dataset} & \textbf{ACC (context-free)} & \textbf{ACC (in-context)} \\
        \midrule
        \textbf{CoNLL2003} & 90.1 & 94.2 \\
        \textbf{AI} & 75.5 & 79.6 \\
        \textbf{Literature} & 78.9 & 83.4 \\
        \textbf{Music} & 84.6 & 88.3 \\
        \textbf{Politics} & 87.4 & 90.8 \\
        \textbf{Science} & 82.2 & 86.5 \\
        \bottomrule
    \end{tabular}}
    \caption{Both base models used are Qwen2.5-7B. The results are obtained in zero-shot scenarios. The used datasets are CoNLL2003 and CrossNER. Accuracy is used in the evaluation of classifiers.}
    \label{context_compare}
\end{table}

\begin{table*}[h!]
\centering
\resizebox{0.8\textwidth}{!}{%
\begin{tabular}{|l|cc|cc|cc|}
\hline
\multirow{2}{*}{\textbf{Model}} & \multicolumn{2}{c|}{\textbf{Base-Supervised}} & \multicolumn{2}{c|}{\textbf{Dynamic-Supervised}} & \multicolumn{2}{c|}{\textbf{Dynamic-Fewshot}} \\
\cline{2-7}
& \textbf{en} & \textbf{zh} & \textbf{en} & \textbf{zh} & \textbf{en} & \textbf{zh} \\
\hline
\texttt{G-1.5B} & 50.2 $\pm$ 1.3 & 44.1 $\pm$ 1.3 & 47.6 $\pm$ 1.5 & 41.2 $\pm$ 1.4 & 36.9 $\pm$ 1.6 & 35.8 $\pm$ 1.6 \\
\texttt{G-7B}   & 55.1 $\pm$ 1.2 & 50.9 $\pm$ 1.2 & 52.3 $\pm$ 1.3 & 48.1 $\pm$ 1.3 & 42.7 $\pm$ 1.4 & 41.1 $\pm$ 1.4 \\
\texttt{G-GPT}  & 62.4 $\pm$ 1.1 & 60.4 $\pm$ 1.2 & 60.6 $\pm$ 1.3 & 58.4 $\pm$ 1.3 & 49.2 $\pm$ 1.4 & 48.9 $\pm$ 1.3 \\
\hline
\texttt{P-1.5B} & 21.6 $\pm$ 1.6 & 20.7 $\pm$ 1.5 & 23.2 $\pm$ 1.7 & 22.7 $\pm$ 1.7 & 20.5 $\pm$ 2.0 & 19.9 $\pm$ 1.8 \\
\texttt{P-7B}   & 41.1 $\pm$ 1.4 & 47.4 $\pm$ 1.4 & 44.3 $\pm$ 1.5 & 40.4 $\pm$ 1.5 & 39.8 $\pm$ 1.8 & 37.8 $\pm$ 1.8 \\
\texttt{P-GPT}  & 49.7 $\pm$ 1.3 & 48.6 $\pm$ 1.3 & 53.0 $\pm$ 1.5 & 52.3 $\pm$ 1.4 & 49.4 $\pm$ 1.8 & 47.4 $\pm$ 1.9 \\
\hline
\texttt{C-1.5B} & 67.6 $\pm$ 1.1 & 64.0 $\pm$ 1.1 & 62.8 $\pm$ 1.2 & 58.9 $\pm$ 1.1 & 49.7 $\pm$ 1.3 & 48.5 $\pm$ 1.4 \\
\texttt{C-7B}   & 73.8 $\pm$ 1.0 & 69.8 $\pm$ 1.0 & 68.2 $\pm$ 1.1 & 64.5 $\pm$ 1.1 & 55.7 $\pm$ 1.2 & 52.9 $\pm$ 1.2 \\
\texttt{C-GPT}  & 77.1 $\pm$ 0.9 & 74.3 $\pm$ 0.9 & 73.1 $\pm$ 1.0 & 68.3 $\pm$ 1.0 & 61.3 $\pm$ 1.1 & 59.7 $\pm$ 1.1 \\
\hline
\end{tabular}}
\caption{F1 Scores with 95\% CI for English and Chinese. This table directly validates the statistical significance of our findings for representative languages. The consistently narrow CIs for CascadeNER highlight its high performance stability.}
\label{tab:ci_results_combined}
\end{table*}
\section{Statistical Significance Analysis}
\label{sec:appendix_ci}

To prove the statistical significance of our results on DynamicNER, we conduct an analysis to calculate 95\% confidence intervals (CI) for our primary findings. This analysis validates whether the observed performance differences between methods, particularly the advantages of CascadeNER, are statistically meaningful.

We employ bootstrap resampling over the test set to compute the 95\% CI for the F1 scores. For a test set of size $N$, we create a bootstrap sample by randomly drawing $N$ instances with replacement. The F1 score is then calculated on this new sample. By repeating this process 500 times, we obtain an empirical distribution of F1 scores. The 95\% confidence interval is then determined by taking the 2.5th and 97.5th percentiles of this distribution. This analysis is performed on English and Chinese as two representative languages. The results are presented in Table~\ref{tab:ci_results_combined}.

The results from our statistical analysis yield two key insights. First, the performance gains of CascadeNER over the baseline methods are confirmed to be statistically significant, as the confidence intervals show clear separation in most cases, particularly for the stronger GPT-4o backbone. Second, CascadeNER consistently exhibits the narrowest confidence intervals across almost all settings. This is especially evident when compared to PromptNER, whose complex prompts lead to higher performance variance. This indicates that by decomposing the NER task into two simpler sub-tasks, CascadeNER not only achieves higher accuracy but also offers a more stable and reliable performance.

\section{LLM-based Methods Comparsion}
\label{extraction}
\begin{figure}[H]
    \centering
    \includegraphics[width=0.48\textwidth]{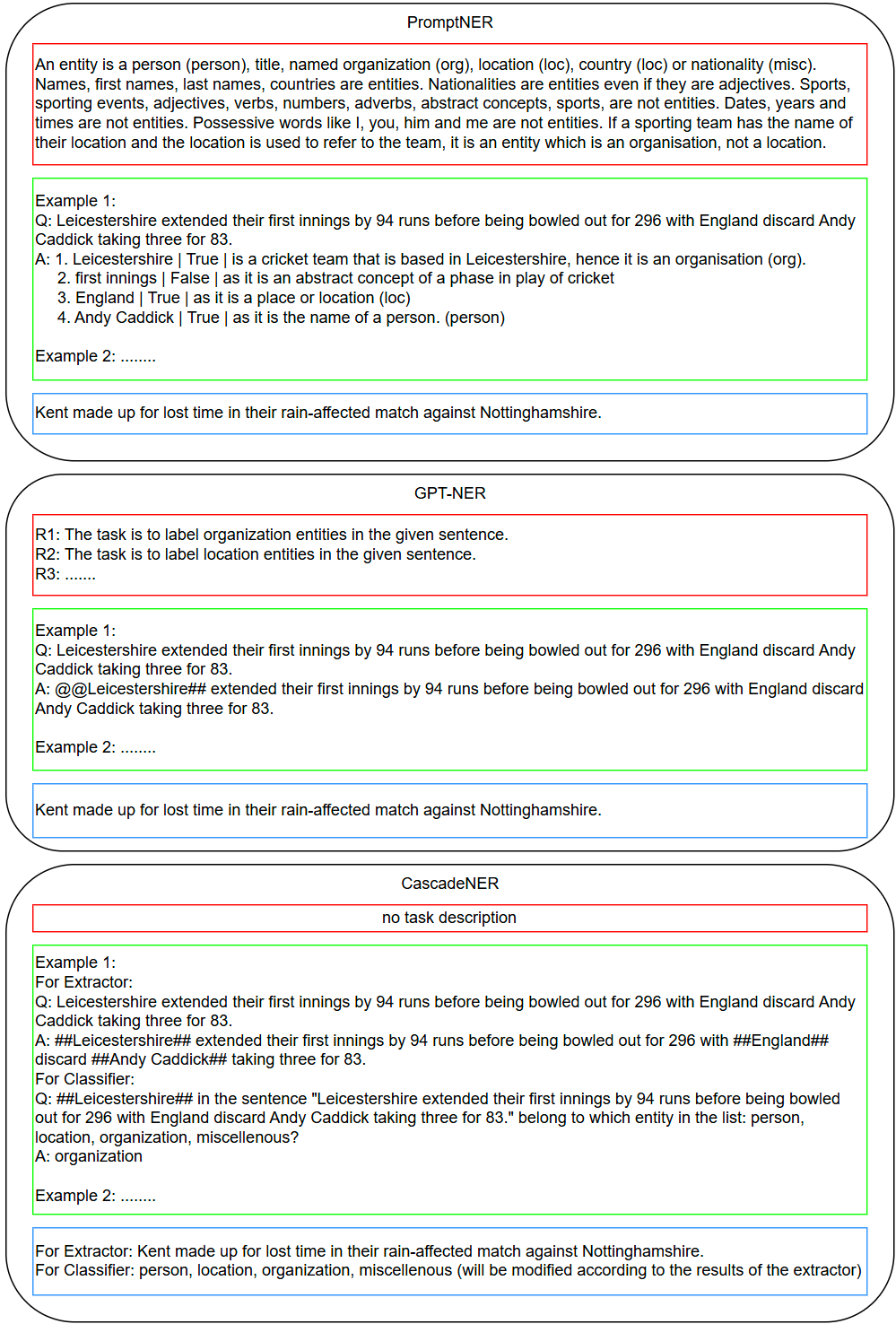}
    \caption{Examples for the three parts of the prompt for each method. The red boxes contain the task description, the green boxes contain few-shot demonstrations, and the blue boxes contain the input sentence.}
    \label{prompt_compare}
\end{figure}

In this section, we compare our prompt with two existing LLM-based baselines, GPT-NER \cite{wang2023gpt} and PromptNER \cite{ashok2023promptner}. These methods are the currently main methods to achieve general NER with LLMs. A breif comparsion is shown in Figure \ref{prompt_compare}.

PromptNER utilizes detailed descriptions of each entity's specific definition and CoT reasoning processes to fully leverage the LLM's logical reasoning abilities. However, like traditional methods, it treats NER as a sequence labeling task, failing to effectively utilize the LLM's global contextual understanding capabilities, making it prone to overlooking important context in complex sentences. Additionally, the task descriptions are overly complex, which not only makes it difficult for lightweight LLMs to correctly execute tasks, but also leads to a higher likelihood of hallucinations in tasks requiring fine-grained classification, such as the fine-grained settings of FewNERD and DynamicNER, as each category's definition requires descriptions. These issues reduce PromptNER's generalization and accuracy, limiting its application.

GPT-NER handles the NER task by determining whether a single entity belongs to a specific category, which leverages the generative capabilities of LLMs and allows for improved attention to the influence of context on entity meaning. Its drawback lies in the fact that it can only process one entity type at a time. This makes the method highly inefficient when dealing with fine-grained categorization, leading to significant resource consumption. Additionally, this method requires multiple judgments for the same entity, introducing the potential for conflicts between different rounds. Unfortunately, GPT-NER does not provide an effective solution for this issue.

CascadeNER divides the NER task into two sub-tasks: extraction and classification, while simplifying the input and output formats and reducing logical complexity. This ensures that even lightweight LLMs with limited capacity to handle complex tasks can still perform the tasks accurately and efficiently. In extraction, CascadeNER leverages the model's generation capabilities by producing sentences with identifiers, treating all entities uniformly, which enhances the model's generalization ability across different languages and domains. Notably, it avoids reliance on word order by consistently using "\#\#" to mark entities, ensuring consistent annotation regardless of whether the language is right-to-left or left-to-right, improving cross-language consistency and adaptability. In classification, our method processes the entire sentence as a whole, better utilizing LLMs' strengths in contextual understanding and semantic modeling. By leveraging the LLM's ability to model long-range dependencies, the model's capacity to handle complex sentence structures is enhanced, avoiding fragmentation of information and improving overall consistency and generalization. However, our method also has limitations. The use of unified identifiers prevents CascadeNER from effectively handling nested NER. We plan to address this by developing a solution that accommodates both multilingual and nested NER tasks in future.

\section{Computational Resource Usage Record}
\label{price}
In Table \ref{cost}, we provide the API costs incurred when testing the complete dynamic version of DynamicNER in few-shot scenarios using three LLM-based methods with GPT-4o, serving as a reference for the computational resources required by these methods. The cost calculation follows OpenAI's official GPT-4o pricing, with input costs at 2.5 USD per 1M tokens and output costs at 10 USD per 1M tokens. The records show that CascadeNER exhibits significant advantages over existing methods in computational resource consumption.
\begin{table}[H]
\centering
\setlength{\tabcolsep}{1mm}
\resizebox{0.48\textwidth}{!}{
\begin{tabular}{lccc}
\toprule
\textbf{} & \textbf{GPT-NER} & \textbf{PromptNER} & \textbf{CascadeNER} \\
\midrule
\textbf{Cost (USD)} & 513.92 & 128.49 & 45.86 \\
\bottomrule
\end{tabular}}
\caption{Cost comparison of three LLM-based methods. The cost is calculated according to OpenAI's official GPT-4o pricing, not the actual cost.}
\label{cost}
\end{table}

\section{Human Annotation Process}
\label{humananno}

\subsection{Manual Annotation}
For the annotators, each language in DynamicNER is annotated by two junior or higher-level students from the corresponding language departments at our universities. Each annotator receives specific training and follows DynamicNER's multi-granularity classification system to ensure consistent and accurate entity annotations across various languages and domains. The annotation process for each language are divided into two parts equally, with each annotator independently handling one part. After the initial annotation, the annotators revise their work based on the review results with the aid of Claude 3.5 Sonnet \cite{Claude2023}. For ambiguous terms or specialized domain terms, the annotators either collaborate with each other or consult experts via personal contact to ensure the accuracy and reliability of the annotation process.

\section{Inter-Annotator Agreement Analysis}
\label{sec:appendix_iaa}

To empirically validate the quality and semantic correctness of DynamicNER, we conduct a Inter-Annotator Agreement (IAA) analysis. We focus on the \texttt{Dynamic Version}, to ensure that the entity labels remain semantically robust and reliable even after undergoing the automated, metric-driven dynamic categorization process.

We select Cohen's Kappa ($\kappa$) as our evaluation metric, as it is a widely recognized statistical measure for evaluating the reliability of judgments between two independent annotators. The process includes two stages. Firstly, we apply our stratified sampling method (detailed in Appendix~\ref{cascadener}) to randomly sample 200 sentences each for English, French, Japanese, and Chinese from the \texttt{Dynamic Version}. These languages are chosen as representative cases for the analysis. The sampled sentences are then given to two annotators who are not involved in the original dataset creation. They perform the annotation in a double-blind setting, where neither annotator was aware of the other's decisions or the original labels. The results of our IAA analysis are presented in Table~\ref{tab:iaa_scores}.

\begin{table}[h]
\centering
\caption{Cohen's Kappa ($\kappa$) scores from the IAA analysis conducted on the dynamic version of DynamicNER.}
\label{tab:iaa_scores}
\begin{tabular}{lc}
\toprule
\textbf{Language} & \textbf{Cohen's Kappa ($\kappa$)} \\
\midrule
English & 0.86 \\
French & 0.82 \\
Chinese & 0.83 \\
Japanese & 0.81 \\
\bottomrule
\end{tabular}
\end{table}

According to established interpretations of the Kappa statistic, the obtained scores indicate substantial agreement. This strong quantitative evidence, derived from a rigorous human-centric audit, directly confirms that the entity labels in our dataset are semantically robust and reliable. It serves as a crucial validation that our automated quality-control pipeline successfully preserves high-quality annotations, thereby addressing the concern about the semantic correctness of the dynamically generated labels.

\section{Finetuning Hyperparameter}
\label{hyperparameter}
To ensure the reproducibility of our experiments, this section provides a detailed record of the hyperparameter configurations used for fine-tuning the lightweight LLMs in CascadeNER.

Our implementation is based on the well-regarded MS-SWIFT framework \cite{swift}, and we employ Low-Rank Adaptation (LoRa) \cite{hu2022lora} for Parameter-Efficient Fine-Tuning (PEFT) \cite{houlsby2019parameter}. All fine-tuning and inference experiments are conducted on a server equipped with 8 NVIDIA RTX 3090 GPUs with 24GB VRAM for each GPU.

\definecolor{LightBlue}{rgb}{0.8, 0.85, 1}
\begin{table}[H]
    \centering
    \setlength{\tabcolsep}{2mm}
    \resizebox{0.4\textwidth}{!}{
    \begin{tabular}{ll}
    \toprule
    \textbf{Hyperparameter} & \textbf{Value} \\
    \midrule
    \rowcolor{LightBlue}
    \multicolumn{2}{l}{\textit{LoRA Parameters}} \\
    Target Modules & ALL \\
    Rank (r) & 8 \\
    Alpha ($\alpha$) & 32 \\
    Dropout & 0.05 \\
    \midrule
    \rowcolor{LightBlue}
    \multicolumn{2}{l}{\textit{Training Parameters}} \\
    Learning Rate & 1e-4 \\
    Optimizer & AdamW \\
    Weight Decay & 0.01 \\
    Max Sequence Length & 4096 \\
    Training Epochs & Max 50 \\
    \quad - Best Epoch (1.5B model) & 25 \\
    \quad - Best Epoch (7B model) & 20 \\
    Effective Batch Size & 32 \\
    \bottomrule
    \end{tabular}
    }
    \caption{Hyperparameters used for fine-tuning CascadeNER models. The 'ALL' setting for Target Modules indicates that LoRA was automatically applied to all suitable linear layers within the model.}
    \label{tab:hyperparams}
\end{table}

The specific parameters for LoRA and the training process are detailed in Table~\ref{tab:hyperparams}. We maintain a consistent effective batch size of 32 for both models to ensure a fair comparison. For the 1.5B model, we use 4 GPUs with a per-device batch size of 4 and 2 gradient accumulation steps. For the 7B model, we use 8 GPUs with a per-device batch size of 1 and 4 gradient accumulation steps. We train each model for a maximum of 50 epochs, saving a checkpoint every 5 epochs and using an early stopping strategy based on the validation set performance. The best-performing checkpoints are then selected for the final evaluation.
\section{Prompt Example for CascadeNER}
\label{prompt_exp}
\begin{figure*}[t]
    \centering
    \includegraphics[width=1\textwidth]{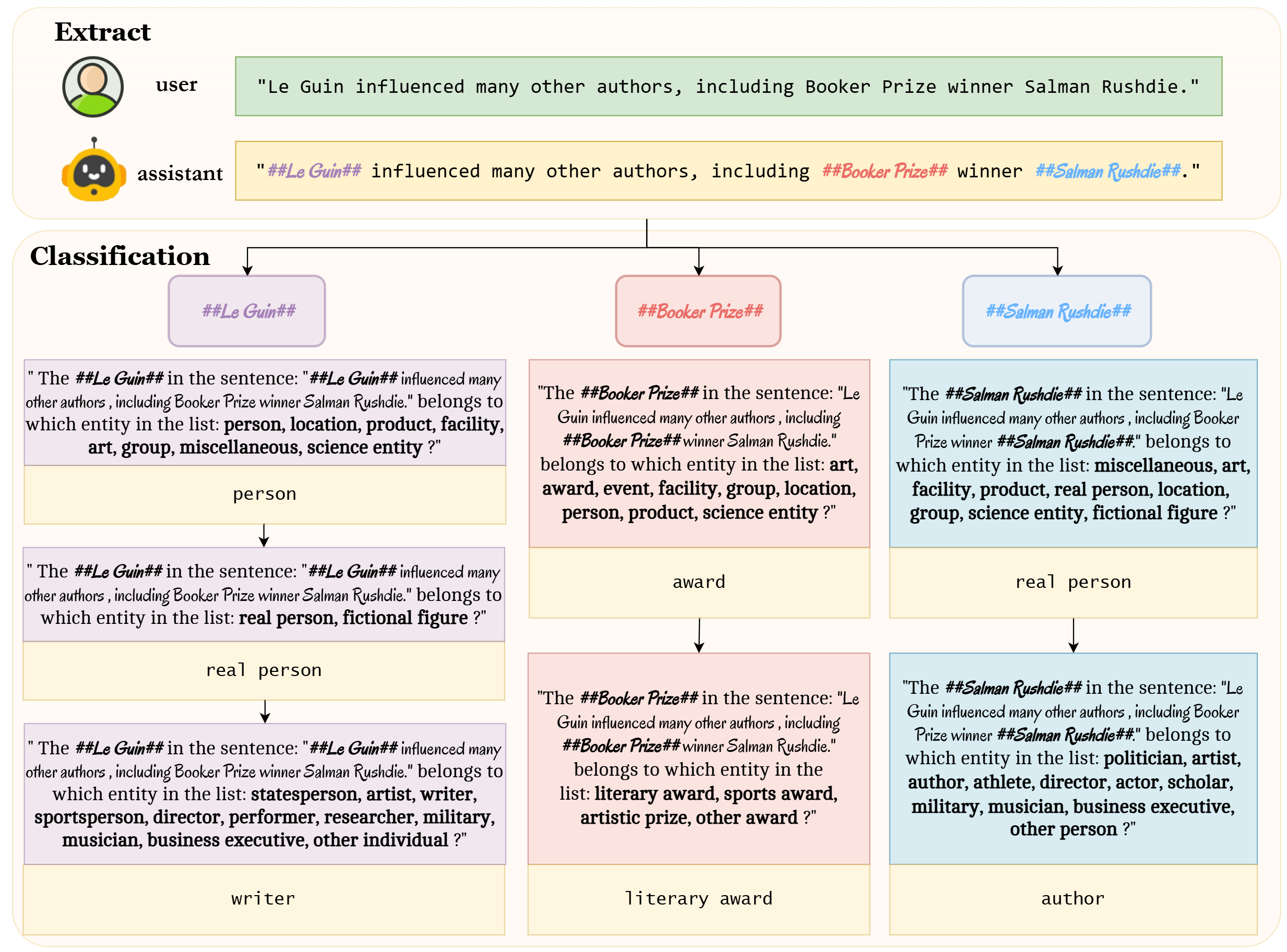}
    \caption{Examples for the complete prompts of CascadeNER.}
    \label{detailed_prompt_example}
\end{figure*}

In this section, we present the examples of prompts used in CascadeNER to provide an intuitive demonstration of the method and to facilitate reproducibility. The example is provided in Figure \ref{detailed_prompt_example}. We provide complete prompts and corresponding json files for training in our github repository.

\section{Data Contamination Statement}
\label{contamination}
Given that LLMs are trained on data from diverse and complex sources, there is a possibility that portions of the evaluation sets may have been encountered during pre-training. However, as prior research \cite{chowdhery2023palm} indicates, contaminated data that has been seen during training does not significantly influence performance. Thus, we consider this issue negligible.
\\
In additional experiments on CascadeNER, we notice another critical data contamination concern: potential corpus overlap between DynamicNER and other benchmark datasets utilizing Wikipedia-derived text, which can reduce evaluation fairness. To mitigate this risk, we implement a rigorous filtering protocol during DynamicNER's annotation phase. After completing the initial manual annotation of the base version, we employ Sentence-BERT to compute semantic cosine similarity between each candidate sentence and existing sentences in reference datasets. Sentences exhibiting similarity scores exceeding 0.8 are excluded from the corpus. New sentences from collected corpus meeting the similarity criteria are then re-annotated following the original annotation workflow. This iterative process continues until all sentences in the base version satisfy the similarity constraints. After this we utilize dynamic categorization to generates the dynamic version. This procedure ensures the reliability and fairness of our test results.

\section{Ethical Statement}
\label{ethical}
When constructing DynamicNER, we strictly adhere to existing ethical guidelines \cite{bender2018data, gebru2021datasheets, hovy2016social}, ensuring that our data sources and processing methods comply with legal and ethical standards while maintaining high-quality annotations. All the text in DynamicNER is sourced from Wikipedia, ensuring no violations of privacy or copyright, as Wikipedia is an open-source platform with user-contributed content from around the world. During data collection and annotation, we balance category distribution to minimize the risk of bias in the model. Furthermore, we maintain transparency by detailing the dataset development process and data partitioning in this paper, ensuring clarity and reproducibility for future research.
\\
In our writing, we use ChatGPT-4o \cite{achiam2023gpt} and Claude 3.5 Sonnet \cite{Claude2023} for assistance.

\section{Detailed Categories of DynamicNER}
\label{detailcategory}
\subsection{Person}

\paragraph{Real Person}
Politician, Artist, Author, Athlete, Director, Actor, Scholar, Military, Musician, Business Executive, Other Person.

\paragraph{Fictional Figure}
Mythological Figure, Other Figure.

\subsection{Location}

\paragraph{Geographical Entity}
Water Body, Mountain, Island, Desert, Other Geographical Entity.

\paragraph{Geo-Political Entity}
Continent, Country, State or Province, City, District, Region, Other GPE.

\paragraph{Address}
Address, Road, Railway, Other Address.

\subsection{Product}

\paragraph{Food}
Beverages, Packaged Foods, Other Food.

\paragraph{Weapon}
Firearms, Biological, Chemical Weapon, Explosives, Cold Weapon, Nuclear, Other Weapon.

\paragraph{Technology}
Software, Website, Electronics, AI, Other Technology.

\paragraph{Vehicle}
Air, Car, Water, Rail, Bike, Other Vehicle.

\paragraph{Other Product}
Clothes, Household, Personal Care, Toys, Musical Instruments, Other Product.

\subsection{Facility}

\paragraph{Public Facility}
Hospital, Library, Park, Landmark, School, Museum, Sports Facility, Other Public Facility.

\paragraph{Commercial Facility}
Hotel, Restaurant, Market/Mall, Theater/Cinema, Bank, Other Commercial Facility.

\paragraph{Transportation Facility}
Airport, Station, Port, Other Transportation Facility.

\paragraph{Production Facility}
Factory, Farm, Mine, Energy, Other Production Facility.

\paragraph{Other Facility}
Residential, Government Facility, Other Facility.

\subsection{Art}

\paragraph{Visual Art}
Painting, Sculpture, Visual Art Genre, Other Visual Art.

\paragraph{Music}
Song, Album, Music Genre, Other Music.

\paragraph{Literature}
Poem, Non-fiction, Fiction, Literature Genre, Other Literature.

\paragraph{Other Art}
Film, Play, Broadcast Program, Game, Other Art.

\subsection{Group}

\paragraph{Social Group}
Ethnic Group, Religious Group, Other Social Group.

\paragraph{Non-commercial Organization}
Educational and Research, Political/Military, Community, Religious Organization, Other Non-commercial Organization.

\paragraph{Commercial Organization}
Sports Team, Band, Company, Media, Other Commercial Organization.

\subsection{Miscellaneous}

\paragraph{Award}
Literary Award, Sports Award, Artistic Award, Other Award.

\paragraph{Event}
Political/Military Event, Sporting Event, Disaster, Business Event, Other Event.

\paragraph{Miscellaneous}
Educational Degree, Tradition, God, Law, Language, Miscellaneous.

\subsection{Science Entity}

\paragraph{Biological}
Protein, Species, Biological Theory, Other Biological Entity.

\paragraph{Chemical}
Element, Compound, Reaction, Chemical Theory, Other Chemical Entity.

\paragraph{Physical}
Physical Phenomenon, Astronomical Object, Physical Theory, Other Physical Entity.

\paragraph{Computer Science}
ProgramLang, Algorithm, Other Computer Science Entity.

\paragraph{Medical}
Disease, Injury, Medication, Symptom, Medical Theory, Other Medical Entity.

\paragraph{Other Scientific Entity}
Discipline, Academic Journal, Conference, Metrics, Other Scientific Entity.

\section{More Categorization Quality Evaluation}
\label{appendix_quality}
\begin{figure*}[t]
    \centering
    \includegraphics[width=1\textwidth]{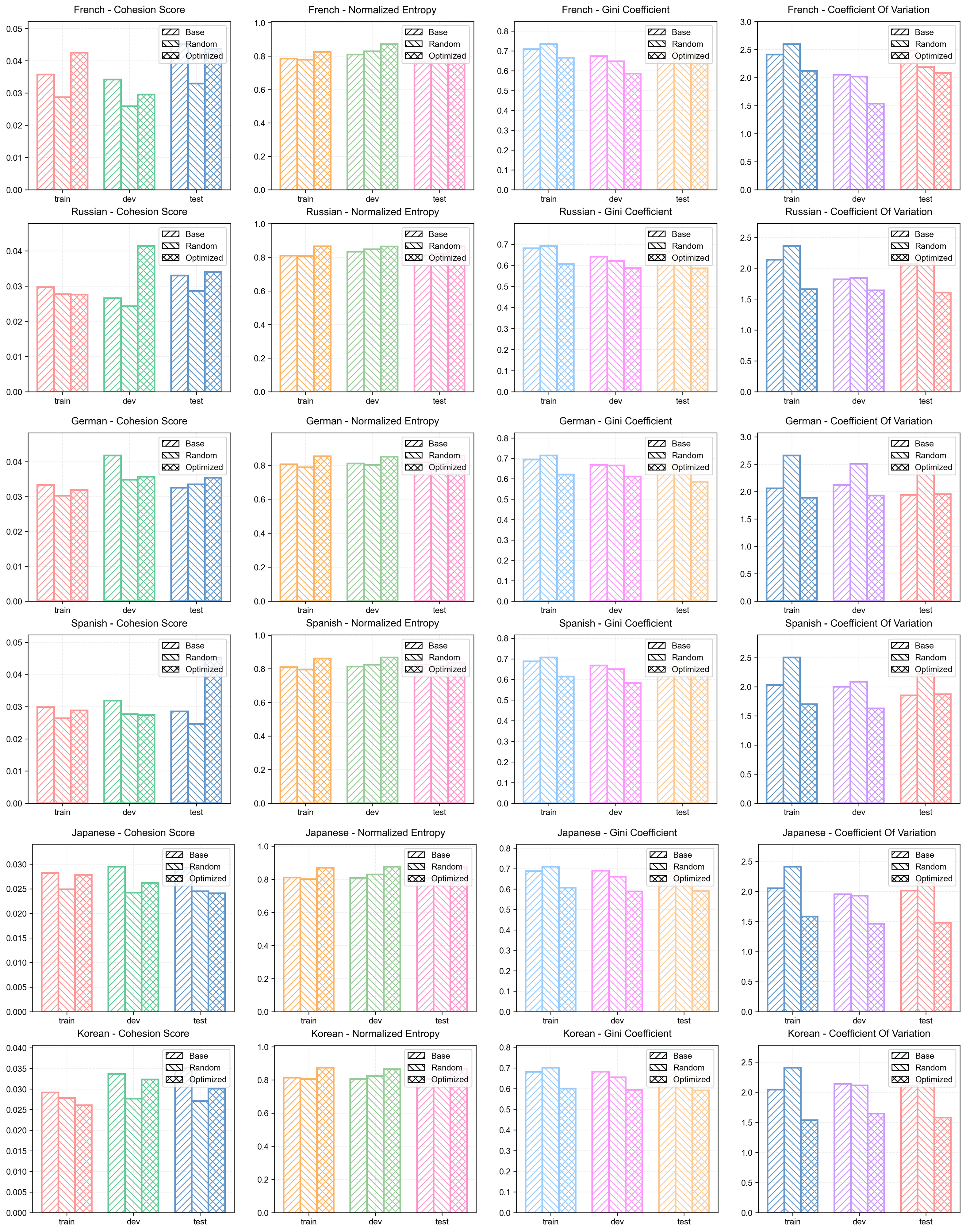}
    \caption{Quantitative categorization metric results for 3 versions DynamicNER.}
    \label{fr-ru}
\end{figure*}

In this section, we display the quantitative results of categorization metrics in Spanish, French, Russian, German, Japanese, and Korean. The results in shown in Figure \ref{fr-ru}. Experimental results demonstrate that our dynamic categorization method maintains or improves dataset quality compared to the base version in all languages.
\end{document}